\documentclass{article}

\PassOptionsToPackage{numbers}{natbib}
\usepackage[final]{nips_2016}

\usepackage[utf8]{inputenc} % allow utf-8 input
\usepackage[T1]{fontenc}    % use 8-bit T1 fonts
\usepackage{hyperref}       % hyperlinks
\usepackage{url}            % simple URL typesetting
\usepackage{booktabs}       % professional-quality tables
\usepackage{amsfonts}       % blackboard math symbols
\usepackage{nicefrac}       % compact symbols for 1/2, etc.
\usepackage{microtype}      % microtypography

\usepackage{verbatim}
\usepackage{pgfplots}
\usepackage{placeins}
\usepackage{appendix}
\usepackage{amsmath}
\usepackage{algorithm}
\usepackage{algpseudocode}

\usepackage{caption}
\usepackage{subcaption}
\usepackage{tikz}
\usetikzlibrary{arrows,arrows.meta,positioning,fit}
\usetikzlibrary{backgrounds,patterns}
\tikzset{
    >=stealth',
    punkt/.style={
           rectangle,
           rounded corners,
           draw=black, very thick,
           minimum width=5em,
           minimum height=2em,
           text centered},
    pil/.style={
           ->,
           thick,
           shorten <=2pt,
           shorten >=2pt},
    box/.style={
        minimum size=1cm,
        draw=black,
        very thick
    },
    bggood/.style={
        minimum size=0.33cm,
        fill=green,
        opacity=0.5
    },
    bgbad/.style={
        minimum size=0.33cm,
        fill=red,
        opacity=0.5,
    }
}

\title{Learning to Avoid Errors in GANs by Manipulating Input Spaces}

\author{
  Alexander B. Jung \\
  TU Dortmund \\
  \texttt{alexander2.jung@tu-dortmund.de} \\
}

\begin{document}
% \nipsfinalcopy is no longer used

\maketitle

\begin{abstract}
  Despite recent advances, large scale visual artifacts are still a common occurrence in images generated by GANs. Previous work has focused on improving the generator's capability to accurately imitate the data distribution $p_{data}$. In this paper, we instead explore methods that enable GANs to actively avoid errors by manipulating the input space. The core idea is to apply small changes to each noise vector in order to shift them away from areas in the input space that tend to result in errors. We derive three different architectures from that idea. The main one of these consists of a simple residual module that leads to significantly less visual artifacts, while only slightly decreasing diversity. The module is trivial to add to existing GANs and costs almost zero computation and memory. Code is available at: \url{https://github.com/aleju/gan-error-avoidance}
\end{abstract}

\section{Introduction}

Recently, Generative Adversarial Networks~\cite{GAN} (GANs) have received quickly growing attention. They are a class of generative methods that use neural networks to map input spaces to output spaces, with the former one being random noise and the latter one usually being images. The mapping is optimized to imitate the probability distribution of real data. Intuitively, a GAN consists of a forger (the generator) and a policeman (the discriminator). The forger is trained to generate good-looking fake images, while the policeman is trained to differentiate between real and fake images. In such a scenario, the forger may chose to focus on producing images with specific features. For instance, when creating human portraits, the forger may notice that he is bad at drawing large noses and unable to improve upon that skill, even after having forged many images with that feature. In such a scenario, the forger could chose to simply avoid drawing large noses altogether and entirely focus on smaller ones. In the context of GANs, the assumption is usually that the generator should be able to automatically show such a behaviour in order to trick the discriminator. However, in practice that is often not the case as large-scale visual artifacts are still common. As the reconstruction results in~\cite{WN} show, even shallow generator architectures have enough capacity to create realistic looking images -- if the input vector is carefully chosen. In this work, we suggest to give the generator some influence over the input space in order to manipulate noise vectors. This allows the model to actively avoid input vectors for which it would generate bad-looking images. Briefly summarized, we allow the model to change the input vector in a way that (a) leads to better images according to the discriminator and (b) keeps the new vectors as close as possible to the old ones.

The contribution of this paper is the introduction of an alternative way to optimize GANs, which is orthogonal to the existing one that focuses on the mapping from noise vector to output. We introduce the idea of learned input space manipulation, a technique that allows the generator to actively avoid making errors. % \footnote{Despite the simplicity of the idea, we are not aware of any other paper explicitly discussing similar thoughts. However, as mentioned in the section "Related Work", some authors might have unintentionally used error avoidance in their architectures.}
Depending on how this is implemented, one can argue that the generator becomes aware of its own skills and especially weaknesses. \\
From this general idea we derive three architectures of varying complexity. The main one of these is based on LIS modules, which reduce the frequency of visual artifacts while being almost free in terms of computation and memory. Time constrained readers are recommended to take a look at figure~\ref{fig:LIS} and optionally read section~\ref{sec:G-LIS} as well as the introduction to section~\ref{sec:proposed_method}.

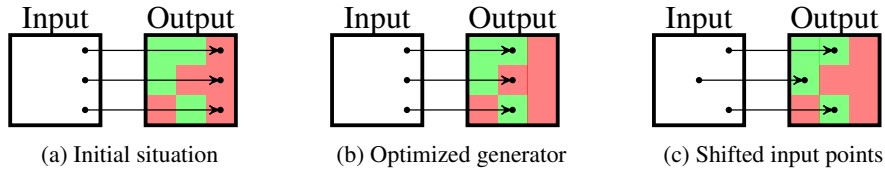
\begin{figure}[t]
\centering
\begin{subfigure}{.3\textwidth}
\centering
\scalebox{1.2}{
\begin{tikzpicture}
    \node[] at (0, 0.66) (input-label) {Input};
    \node[] at (1.5, 0.66) (input-label) {Output};

    % left column
    \node[bggood] at (1.5-0.33, 0.33) (output) {};
    \node[bggood] at (1.5-0.33, 0) (output) {};
    \node[bgbad] at (1.5-0.33, -0.33) (output) {};
    
    % middle column
    \node[bggood] at (1.5-0, 0.33) (output) {};
    \node[bgbad] at (1.5-0, 0) (output) {};
    \node[bggood] at (1.5-0, -0.33) (output) {};
    
    % right column
    \node[bgbad] at (1.5+0.33, 0.33) (output) {};
    \node[bgbad] at (1.5+0.33, 0) (output) {};
    \node[bgbad] at (1.5+0.33, -0.33) (output) {};
    
    \node[box] at (0, 0) (input) {};
    \node[box] at (1.5, 0) (output) {};
    
    % input points
    \fill (0+0.33, 0.33) circle [radius=1pt];
    \fill (0+0.33, 0) circle [radius=1pt];
    \fill (0+0.33, -0.33) circle [radius=1pt];
    
    % output points
    \fill (1.5+0.33, 0.33) circle [radius=1pt];
    \fill (1.5+0.33, 0) circle [radius=1pt];
    \fill (1.5+0.33, -0.33) circle [radius=1pt];
    
    \draw[->] (0+0.33, 0.33) -- (1.5+0.33, 0.33);
    \draw[->] (0+0.33, 0) -- (1.5+0.33, 0);
    \draw[->] (0+0.33, -0.33) -- (1.5+0.33, -0.33);
\end{tikzpicture}
}
\label{fig:overview-initial}
\caption{Initial situation}
\end{subfigure}
\begin{subfigure}{.3\textwidth}
\centering
\scalebox{1.2}{
\begin{tikzpicture}
    \node[] at (0, 0.66) (input-label) {Input};
    \node[] at (1.5, 0.66) (input-label) {Output};

    % left column
    \node[bggood] at (1.5-0.33, 0.33) (output) {};
    \node[bggood] at (1.5-0.33, 0) (output) {};
    \node[bgbad] at (1.5-0.33, -0.33) (output) {};
    
    % middle column
    \node[bggood] at (1.5-0, 0.33) (output) {};
    \node[bgbad] at (1.5-0, 0) (output) {};
    \node[bggood] at (1.5-0, -0.33) (output) {};
    
    % right column
    \node[bgbad] at (1.5+0.33, 0.33) (output) {};
    \node[bgbad] at (1.5+0.33, 0) (output) {};
    \node[bgbad] at (1.5+0.33, -0.33) (output) {};
    
    \node[box] at (0, 0) (input) {};
    \node[box] at (1.5, 0) (output) {};
    
    % input points
    \fill (0+0.33, 0.33)  circle [radius=1pt];
    \fill (0+0.33, 0.00)  circle [radius=1pt];
    \fill (0+0.33, -0.33) circle [radius=1pt];
    
    % output points
    \fill (1.5+0, 0.33)  circle [radius=1pt];
    \fill (1.5+0, 0.00)  circle [radius=1pt];
    \fill (1.5+0, -0.33) circle [radius=1pt];
    
    \draw[->] (0+0.33, 0.33)  -- (1.5+0, 0.33);
    \draw[->] (0+0.33, 0.00)  -- (1.5+0, 0);
    \draw[->] (0+0.33, -0.33) -- (1.5+0, -0.33);
\end{tikzpicture}
}
\label{fig:overview-generator}
\caption{Optimized generator}
\end{subfigure}
\begin{subfigure}{.3\textwidth}
\centering
\scalebox{1.2}{
\begin{tikzpicture}
    \node[] at (0, 0.66) (input-label) {Input};
    \node[] at (1.5, 0.66) (input-label) {Output};

    % left column
    \node[bggood] at (1.5-0.33, 0.33) (output) {};
    \node[bggood] at (1.5-0.33, 0) (output) {};
    \node[bgbad] at (1.5-0.33, -0.33) (output) {};
    
    % middle column
    \node[bggood] at (1.5-0, 0.33) (output) {};
    \node[bgbad] at (1.5-0, 0) (output) {};
    \node[bggood] at (1.5-0, -0.33) (output) {};
    
    % right column
    \node[bgbad] at (1.5+0.33, 0.33) (output) {};
    \node[bgbad] at (1.5+0.33, 0) (output) {};
    \node[bgbad] at (1.5+0.33, -0.33) (output) {};
    
    \node[box] at (0, 0) (input) {};
    \node[box] at (1.5, 0) (output) {};
    
    % input points
    \fill (0+0.33, 0.33)  circle [radius=1pt];
    \fill (0+0.00, 0.00)  circle [radius=1pt];
    \fill (0+0.33, -0.33) circle [radius=1pt];
    
    % output points
    \fill (1.5+0, 0.33)  circle [radius=1pt];
    \fill (1.5-0.33, 0.00)  circle [radius=1pt];
    \fill (1.5+0, -0.33) circle [radius=1pt];
    
    \draw[->] (0+0.33, 0.33)  -- (1.5+0, 0.33);
    \draw[->] (0+0.00, 0.00)  -- (1.5-0.33, 0);
    \draw[->] (0+0.33, -0.33) -- (1.5+0, -0.33);
\end{tikzpicture}
}
\label{fig:overview-lis}
\caption{Shifted input points}
\end{subfigure}

\caption{
    Overview of the suggested idea. For simplicity, the generator is here assumed to be of limited capacity, forcing it to apply the exactly same operation to every input point.
    (a) Initial situation. A generator maps points from an input space (usually random noise) to a high-dimensional output space (usually images). The initial output of the generator is of low quality (e.~g. broken images), indicated by the red color of each output point's location.
    (b) Situation after optimizing the generator. Two of the input points are now mapped to high quality outputs, while one remains broken. Due the generator's inherent limitation, it is unable to improve further, independent of the training time.
    (c) One of the input points is slightly shifted, which enables the generator to produce high quality outputs for all input points. Such an operation would be impossible in most other machine learning scenarios, but in GANs it is a viable option. Note that as only one input point was moved slightly, the expected loss in output diversity is minimal. The model can be forced to move input points as little as possible using a similarity constraint.
}
\label{fig:overview}
\end{figure}

\section{Proposed Method}
\label{sec:proposed_method}

In GANs, the training procedure follows a min-max game according to
\begin{equation}
    \min_{G} \max_{D} V(D,G) = \mathbb{E}_{\mathbf{x}{\sim}p_{data}}\left[\mathrm{log}\left(D\left(\mathbf{x}\right)\right)\right] + \mathbb{E}_{\mathbf{z}{\sim}p_z}\left[\mathrm{log}\left(1-D\left(G\left(\mathbf{z}\right)\right)\right)\right],
\end{equation}
where G is the generator, D the discriminator and $\mathbf{z}{\in}\mathbb{R}^{N_z}$ is a noise vector representing a single point in the input space. Usually, $\mathbf{z}$'s components are independently sampled, either from the standard normal distribution $\mathcal{N}(0,1)$ or from a uniform distribution with interval $[-1,1]$. Our target is to give the model control over the input space. We achieve this by allowing it to change each input vector $\mathbf{z}$ into a new vector $\mathbf{z}'$. In the simplest case, this could be done using a fully connected layer. More specific methods are discussed later, in sections~\ref{sec:R-separate}, \ref{sec:R-iterative} and~\ref{sec:G-LIS}. A crucial trait that all three methods share is the use of a similarity constraint, which forces each $\mathbf{z}'$ to be close to $\mathbf{z}$. We use a mean squared error
\begin{equation}
    \mathcal{L}_R = \frac{1}{N_z} \sum_{i=1}^{N_z} \left(\mathbf{z}_i - \mathbf{z}_{i}'\right)^2
    \label{eq:LR}
\end{equation}
to achieve this.
Applying such a constraint is sensible from a theoretical perspective, as our main goal is to avoid errors, which we expect to not be the standard case. Hence, most input vectors should not need to be significantly changed -- and it is desired to minimize changes to the input space as any such manipulation might negatively affect the output diversity. It is also empirically sound to use a similarity constraint, as not doing so leads to mode collapse problems (see section~\ref{sec:experiments}).

In the following sections we introduce three different architectures. We start with a simple one (\textit{R-separate}), extend it to a smarter but slower architecture (\textit{R-iterative}), which is then collapsed to a model that is both smart and fast (\textit{G-LIS}).

\subsection{R-separate}
\label{sec:R-separate}

\begin{figure}[t]
\centering
\begin{subfigure}{.45\textwidth}
\centering
\begin{tikzpicture}[
    bgnode/.style 2 args={
        rectangle,
        draw,
        inner sep=0pt,
        fit=(#1) (#2)}
]
    %\draw[style=help lines] (-4,2) grid (4,8);

    \node[punkt, dashed] (z) {$\mathbf{z}$};
    \node[punkt, inner sep=5pt,below=0.5cm of z] (G) {G};
    \node[punkt, dashed, inner sep=5pt,below=0.5cm of G] (xf) {$\mathbf{x}_{fake}$};
    \node[punkt, inner sep=5pt,below=0.5cm of xf] (D) {D};
    \node[punkt, dashed, inner sep=5pt,left=0.5cm of D] (xr) {$\mathbf{x}_{real}$};
    \node[punkt, dashed, inner sep=5pt,below=0.5cm of D] (rating) {$rating$};
    \node[punkt, inner sep=5pt,left=0.5cm of G] (R) {R};
    
    \coordinate[above=1.25cm of R] (R-rect-tl-dummy) {};
    \coordinate[left=1.9cm of R-rect-tl-dummy] (R-rect-tl) {};
    \coordinate[below=1.2cm of R] (R-rect-br-dummy) {};
    \coordinate[right=1.1cm of R-rect-br-dummy] (R-rect-br) {};
    \begin{scope}[on background layer]
        \node[bgnode={R-rect-br}{R-rect-tl},fill=blue!10!white] {};
    \end{scope}
    %\coordinate[left=0.9cm of R] (R-text-bl-dummy) {};
    %\coordinate[below=0.9cm of R-text-br-dummy] (R-text-bl) {};
    \node[below left=0.3cm and -1.0cm of R,text width=1.75cm] (R-text) {\small trained once at the end};
    
    \node[right=0.5cm of D] (gradient-D) {};
    \node[right=0.5cm of z] (gradient-z-right) {};
    \node[above right=0.5cm of z] (gradient-z-top-right) {};
    \node[above=0.5cm of z] (gradient-z-top) {};
    \node[above left=0.5cm of z] (gradient-z-top-left) {};
    \node[left=0.5cm of R] (gradient-R-left) {};
    
    \draw [->] (z) -- (G);
    \draw [->] (G) -- (xf);
    \draw [->] (xf) -- (D);
    \draw [->] (xr) -- (D);
    \draw [->] (D) -- (rating);
    \draw [->, rounded corners=0.1cm, to path={-| (\tikztotarget)}] (xf) edge (R);
    \draw [->, rounded corners=0.1cm, to path={|- (\tikztotarget)}] (R) edge (z);
    
    %\draw [->] (gradient-D) -- (gradient-z-right) -- (gradient-z-top);
    %\draw[->, to path={-| (\tikztotarget)}]
    %    (gradient-D) edge (gradient-z-right)
    %    (gradient-z-right) edge (gradient-z-top-right)
    %    (gradient-z-top-right) edge (gradient-z-top);
    \draw[-{Triangle}]
        (gradient-D) edge[double] (gradient-z-right);
    \draw[-{Triangle[scale=1]}, to path={-| (\tikztotarget)}]
        (gradient-z-top) edge[double] (gradient-R-left);
\end{tikzpicture}
\caption{R-separate}
\label{fig:R-separate}
\end{subfigure}%
\begin{subfigure}{.45\textwidth}
\begin{tikzpicture}
    \node[punkt, dashed] (z) {$\mathbf{z}$};
    \node[punkt, inner sep=5pt,below=0.5cm of z] (G) {G};
    \node[punkt, dashed, inner sep=5pt,below=0.5cm of G] (xf) {$\mathbf{x}_{fake}$};
    \node[punkt, inner sep=5pt,below=0.5cm of xf] (D) {D};
    \node[punkt, dashed, inner sep=5pt,left=0.5cm of D] (xr) {$\mathbf{x}_{real}$};
    \node[punkt, dashed, inner sep=5pt,below=0.5cm of D] (rating) {$rating$};
    \node[punkt, inner sep=5pt,left=0.5cm of G] (R) {R};
    
    \node[left=1.7cm of R] (horizontalspacer) {};
    
    \coordinate[right=0.5cm of D] (gradient-D) {};
    \coordinate[right=0.5cm of z] (gradient-z-right) {};
    \coordinate[above right=0.5cm of z] (gradient-z-top-right) {};
    \coordinate[above=0.5cm of z] (gradient-z-top) {};
    \coordinate[above left=0.5cm of z] (gradient-z-top-left) {};
    \coordinate[left=0.5cm of R] (gradient-R-left) {};
    
    \draw [->] (z) -- (G);
    \draw [->] (G) -- (xf);
    \draw [->] (xf) -- (D);
    \draw [->] (xr) -- (D);
    \draw [->] (D) -- (rating);
    \draw [->, rounded corners=0.1cm, to path={-| (\tikztotarget)}] (xf) edge (R);
    \draw [->, rounded corners=0.1cm, to path={|- (\tikztotarget)}] (R) edge (z);
    
    %\draw [->] (gradient-D) -- (gradient-z-right) -- (gradient-z-top);
    %\draw[->, to path={-| (\tikztotarget)}]
    %    (gradient-D) edge (gradient-z-right)
    %    (gradient-z-right) edge (gradient-z-top-right)
    %    (gradient-z-top-right) edge (gradient-z-top);
    \draw[-, to path={|- (\tikztotarget)}]
        (gradient-D) edge[double] (gradient-z-top);
        %(gradient-z-right) edge[double] (gradient-z-top);
    
    %\draw[-{Triangle[scale=1]}, to path={-| (\tikztotarget)}]
    \draw[-, to path={-| (\tikztotarget)}]
        (gradient-z-top) edge[-{Triangle[scale=1]},double] (gradient-R-left);
\end{tikzpicture}
\caption{R-iterative}
\label{fig:R-iterative}
\end{subfigure}
\caption{
    Two of the three suggested architectures to avoid errors in GANs by learning better input spaces. Thick arrows indicate gradient flow.
    Both architectures introduce a "reverser" (R), which projects generated images back onto their noise vectors.
    (a) In \textit{R-separate} the reverser is trained once, after the training of the generator (G) and discriminator (D) is finished. Images can then be generated using $G(R(G(\mathbf{z})))$, which in some cases improves image quality due to a regression to the mean effect. Gradients do not flow from D to R, only a mean squared error between $\mathbf{z}$ and $\mathbf{z}'$ (R's prediction of $\mathbf{z}$) is applied.
    (b) In \textit{R-iterative} the reverser is trained end-to-end in conjunction with G and D. After the first execution of G, the combination of R and G can be applied many times, leading to iterative improvements of the input space and therefore better images. R is trained to both trick D and predict $\mathbf{z}$ accurately.
}
\label{fig:R-models}
\end{figure}
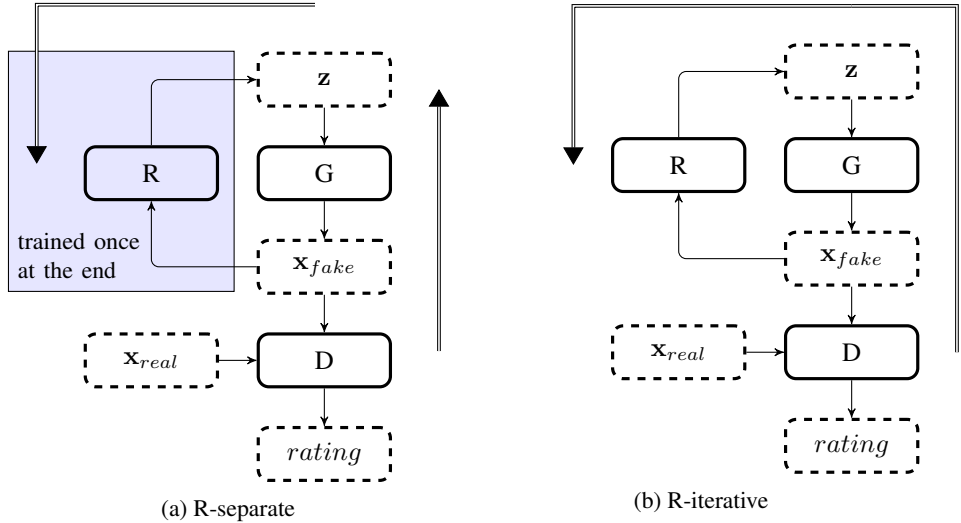
%
% -----------------------
%
\begin{figure}[t]
\centering
\begin{subfigure}{.45\textwidth}
\centering
\begin{tikzpicture}
    \node[punkt, dashed,text width=3cm] (z) {$\mathbf{z}$};
    \node[punkt, inner sep=5pt,below=0.5cm of z,text width=3cm] (lis) {$1..N$ LIS modules};
    \node[punkt, inner sep=5pt,below=0.5cm of lis,text width=3cm] (fc) {Fully Connected};
    \node[punkt, inner sep=5pt,below=0.5cm of fc,text width=3cm] (convs) {$1..N$ Convolution + Upscaling};
    \node[punkt, dashed, inner sep=5pt,below=0.5cm of convs,text width=3cm] (xf) {$\mathbf{x}_{fake}$};
    
    \draw [->] (z) -- (lis);
    \draw [->] (lis) -- (fc);
    \draw [->] (fc) -- (convs);
    \draw [->] (convs) -- (xf);
\end{tikzpicture}
\caption{G-LIS}
\label{fig:G-LIS}
\end{subfigure}%
\begin{subfigure}{.45\textwidth}
\begin{tikzpicture}
    \node[punkt, dashed,text width=1cm] (z) {$\mathbf{z}$};
    \coordinate[below=0.5cm of z] (zbelow) {};
    
    \node[punkt, inner sep=5pt,below right=0.5cm and 1cm of zbelow,text width=3cm] (fc1) {Fully Connected + Normalization + Activation};
    \node[punkt, inner sep=5pt,below=0.5cm of fc1,text width=3cm] (fc2) {Fully Connected};
    \node[punkt, inner sep=5pt,below=4cm of z,text width=1cm] (add) {+};
    \node[punkt, dashed, inner sep=5pt,below=0.5cm of add,text width=1cm] (z2) {$\mathbf{z}'$};
    
    \node[left=0.75cm of z] (horizontalspacer) {};
    
    \draw [->] (z) -- (add);
    \draw [-,rounded corners=0.1cm, to path={-| (\tikztotarget)}] (zbelow) edge[->] (fc1);
    \draw [->] (fc1) -- (fc2);
    \draw [->,rounded corners=0.1cm, to path={|- (\tikztotarget)}] (fc2) edge (add);
    \draw [->] (add) -- (z2);
    
    \coordinate[left=0.5cm of z] (zleft) {};
    \coordinate[left=0.5cm of z2] (z2left) {};
    \draw [<-,dashed] (z2) -- (z2left);
    \draw [-,dashed] (z2left) -- (zleft);
    \draw [->,dashed] (zleft) -- (z);
    \node [rotate=90, below left=0.35cm and 1.1cm of z] {similarity constraint $\mathcal{L}_R$ (MSE)};
\end{tikzpicture}
\caption{LIS module}
\label{fig:LIS_module}
\end{subfigure}
\caption{
    (a) Overview of the generator's architecture in \textit{G-LIS}. Compared to most other \mbox{DCGAN-like}~\cite{DCGAN} architectures, one or more LIS modules are added. These change the input vector $\mathbf{z}$ into a new vector $\mathbf{z}'$ that similar to $\mathbf{z}$, but results in images of higher quality.
    (b) A single LIS ("learned input space") module. It performs an elementwise addition, similar to the technique used in residual networks~\protect\cite{ResNet}. Its goal is to move $\mathbf{z}$ away from locations in the input space that tend to result in visual artifacts. In most cases, only small changes are expected to be necessary for that, which is why the module's output $\mathbf{z}'$ is constrained to be close to $\mathbf{z}$ via a mean squared error. Applying that loss is also crucial in order to avoid mode collapse problems.
}
\label{fig:LIS}
\end{figure}
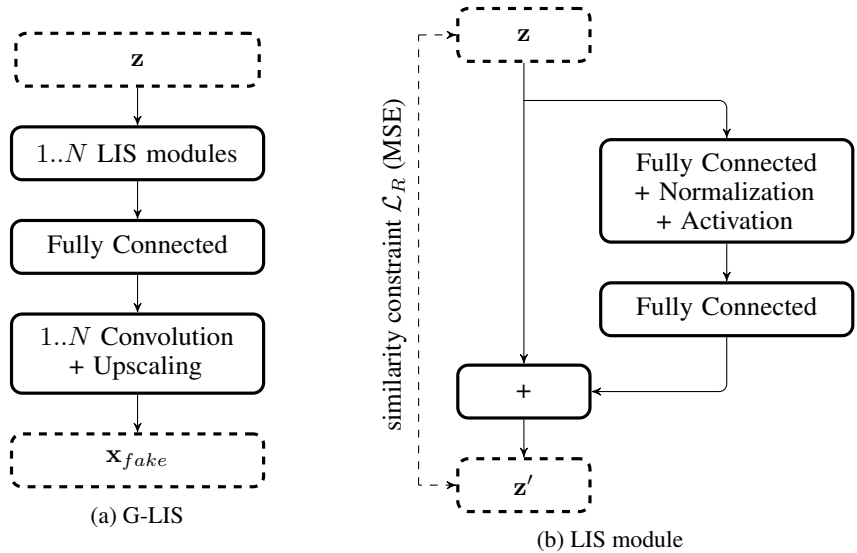
We introduce a new component -- additionally to G and D -- called the "reverser"~(R). In this architecture, R is optimized separately, after the training of G and D is finished (see figure~\ref{fig:R-separate}). It receives images and projects them onto vectors $\mathbf{z}'$. The used loss function is the previously discussed similarity constraint, i.~e. a mean squared error (see equation~\ref{eq:LR}). Due to the loss, R is trained to recover input noise vectors from images generated by G. In order to repair errors, each recovered vector $\mathbf{z}'$ is fed back into G. Hence, the final images are generated by applying $G(R(G(\mathbf{z})))$. That also means that in this architecture, R is not trained on D and as such cannot be expected to have any understanding of G's skills. It still tends to repair image errors due to a regression to the mean effect. Intuitively, assume that each image generated by G is made up of different components (e.~g. noses of varying shapes). If G works mostly correct, then each such component will usually look real, but may occasionally contain artifacts. If R's capacity is limited, it will focus on accurately mapping good-looking components to $\mathbf{z}'$-vectors, as these make up most of the examples. Incorrectly generated components will be mapped by R to $\mathbf{z}'$-vectors of components that both look similar and contain few or no artifacts. The exact algorithm to train \textit{R-separate} is listed in Appendix~\ref{sec:appendix-algorithms}.

\subsection{R-iterative}
\label{sec:R-iterative}
In the previous architecture \textit{R-separate}, R was trained once at the end. Additionally, each image was reversed to $\mathbf{z}'$ one single time. In \textit{R-iterative} (outlined in figure~\ref{fig:R-iterative}), we remove both of these limitations. We train R end-to-end with G and D. The loss is modified to include both $\mathcal{L}_R$ as well as D's ratings. This leads to R choosing $\mathbf{z}'$ in a way that leads to good-looking images according to D (i.~e. less artifacts), which implies that R must have some understanding of G's skills. Due to the similarity constraint $\mathcal{L}_R$, it also leads to each $\mathbf{z}'$ being close to $\mathbf{z}$. To balance the two losses, $\mathcal{L}_R$ is multiplied by $\lambda_R$, which controls the strength of the similarity constraint. Higher values for $\lambda_R$ give R less leeway in finding vectors that result in good-looking images, making error correction harder, but also decreasing the likelihood of mode collapse problems. \\
In contrast to \textit{R-separate}, which was executed one single time per image, \textit{R-iterative} is iteratively applied $N_R$ times, resulting in $1+N_R$ generated images. For instance, the third iteration's image would be generated by $G(R(G(R(G(\mathbf{z})))))$. This allows the model to gradually improve $\mathbf{z}'$. As the assumption is that the last iteration of R resembles the best suited input space, we try to focus the training on that iteration. To achieve this, we start training at the $i$-th iteration with probability $\frac{1+i}{1+N_R}$. If that probability does not hit, we merely generate images and new input vectors $\mathbf{z}'$, but do not train G and R on them. In order to avoid mode collapse problems, the vector $\mathbf{z}$ in $\mathcal{L}_R$ is always the first input vector, not the one from the previous iteration. This choice gives the later iterations less leeway in changing $\mathbf{z}$, as the previous iterations have already altered it. To compensate for that, we exponentiate $\lambda_R$ at the $i$-th iteration with $1+i$, leading to $\lambda_{R}^{1+i}$, which weakens the similarity constraint. The exact algorithm to train \textit{R-iterative} can be found in Appendix~\ref{sec:appendix-algorithms}.
%Errors are not backpropagated beyond the inputs of each R.

\subsection{G-LIS and LIS modules}
\label{sec:G-LIS}
The previously described model \textit{R-iterative} is rather slow to train, as G and R have to be executed several times per batch. As an alternative, we introduce \textit{G-LIS}, which is the collapsed version of \textit{R-iterative}. \textit{G-LIS} optimizes input vectors and generates images in a single forward-pass. It operates directly on the input vectors, instead of performing the costly projection from images to $\mathbf{z}'$. To do this, $N_R$ LIS ("learned input space") modules are used, which resemble $N_R$ iterations of G and R. Each such module contains two fully connected layers and generates an output vector with $N_z$ components that is added to $\mathbf{z}\in\mathbb{R}^{N_z}$ in a residual way. The architecture of \textit{G-LIS} and LIS modules is shown in figure~\ref{fig:LIS}. Using $N_R$ LIS modules also leads to $N_R$ new input vectors per batch, $\mathbf{z}_1', \mathbf{z}_2', \dots, \mathbf{z}_{N_R}'$. For each of these, we apply $\mathcal{L}_R$, forcing them to be close to the initial~$\mathbf{z}$. Similar to \textit{R-iterative}, we relax the similarity constraint for the $i$-th module by multiplying the mean squared error with~$\lambda_{R}^{1+i}$. The loss that G achieved on D is backpropagated through all executed modules. As each LIS module only contains small fully connected layers, their execution is almost free. This allows to always train on the last LIS module, comparable to the last iteration of \textit{R-iterative}. However, we additionally train explicitly on earlier modules by stopping the input vector manipulation batch-wise before the $i$-th module with probability $0.5^{N_R-i}$. This has two advantages. First, it forces the generator to map all points in the original input space to images, which we expect to decrease the probability of mode collapse problems. Second, it allows us to generate images for the outputs of each LIS module. These images can be used to gain insight into the changes caused by each module, which is here used in the experiments section.

\section{Experiments}
\label{sec:experiments}
\begin{figure}[t]
    \includegraphics[width=1\textwidth]{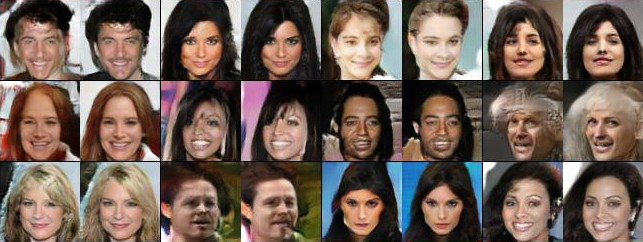}
    \caption{
        Example of \textit{R-separate} applied to a standard generator (without LIS modules), which was only trained for 50k batches and therefore still produced significant artifacts. \textit{R-separate} was trained for 2.5k batches with spatial dropout~\protect\cite{SpatialDropout} of 10\% between convolutional layers. Each of the shown pairs consists of the image before ($G(\mathbf{z})$) and after ($G(R(G(\mathbf{z})))$) applying \textit{R-separate}. The process tends to repair small local artifacts.
    }
    \label{img:r-separate-before-after}
\end{figure}
We conduct experiments for all three suggested architectures on the CelebA dataset~\cite{CelebA}, which contains around 200k images of human faces. We focus on DCGAN-like~\cite{DCGAN} models (i.~e. no Wasserstein distance~\cite{WGAN}) using the recently suggested weight normalization for GANs~\cite{WN}. All images are generated at a resolution of $80\times80$. The noise vectors consist of $256$ components, sampled independently from a standard normal distribution $\mathcal{N}(0,1)$. Our models have mostly identical architectures as in~\cite{WN}. The generator consists of one fully connected layer and four fractionally strided convolutions, starting at 256 filters and ending in 64. The discriminator uses five strided convolutions, starting at 64 and ending in 512 filters. The reverser uses the same architecture as the discriminator, with half as many filters per layer and ending in an output of $256$ values. As suggested in~\cite{WN}, we pair the weight normalization with TPReLUs in all models, which are PReLUs~\cite{PReLUs} with learned thresholds. All model architectures are detailed in Appendix~\ref{sec:appendix-architectures}. We train our models using RMSprop~\cite{RMSProp} with a batch size of~32.

\subsection{R-separate}
\label{sec:experiments-R-separate}
In order to test \textit{R-separate}, we first train a normal pair of G and D for 50k batches at a learning rate of $0.0005$. After that, we train R with 10\% spatial dropout~\cite{SpatialDropout} for up to 50k batches. We observe that R's ability to repair artifacts seems to be strongest during the early batches and decreases over time. This is expected, as the technique is based on a regression to the mean. Longer training leads to R being able to spot more and more details, moving away from the desired mean. Figure~\ref{img:r-separate-before-after} shows example results before and after applying R (at training batch 2.5k). In some cases, we see a decrease in image quality. Usually however, the images are improved or remain stable. Most (non-artifact) small-scale features are kept unchanged. Despite the use of a mean squared error, we do not see an increase in blurriness. Larger example images are listed in Appendix~\ref{sec:appendix-r-separate}.
% and regularly measure the loss of of the discriminator on the old and new images. Figure [TODO] shows D's loss over time, which indicates that the quality of the images is improved, though care has to be taken regarding the optimal number of training batches for R.

\subsection{R-iterative}
\label{sec:experiments-R-iterative}
\begin{figure}[t]
    \begin{subfigure}{1\textwidth}
        \centering
        \includegraphics[width=1\textwidth]{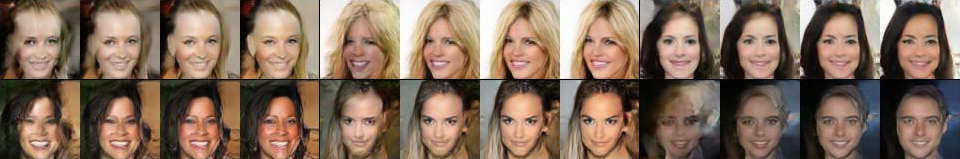}
        \caption{R-iterative}
        \label{img:r-iterative-chains}
    \end{subfigure}
    \begin{subfigure}{1\textwidth}
        \centering
        \includegraphics[width=1\textwidth]{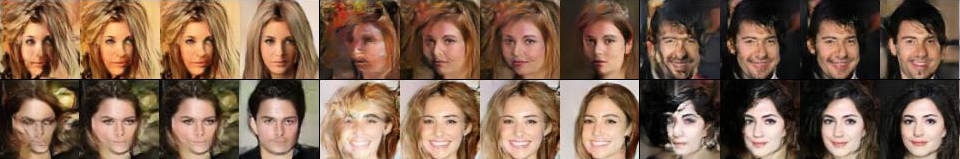}
        \caption{G-LIS with 3 LIS modules}
        \label{img:glis-chains}
    \end{subfigure}
    \caption[R-iterative/G-LIS results]{
        Images generated by (a) \textit{R-iterative} with three iterations~\protect\footnote{When comparing these results to others, keep in mind that G in this experiment is trained up to four times per batch, as there are three iterations of R and G (and one further execution of G at the very start of each batch). As described in section~\protect\ref{sec:R-iterative}, the training is started per batch at the $i$-th iteration with probability~$\frac{1+i}{1+N_R}$, resulting in an average of $\sim2.8$ parameter updates of the generator per batch.} and (b) \textit{G-LIS} with three LIS modules. The improvements made by \textit{R-iterative} are continuous and similar to the ones of recurrent networks, even though it only alters the input vector and refeeds it into the same generator. In the case of \textit{G-LIS}, the improvements are not as continuous. The first and especially the third module seem to have significant effects, while the second one changes the input vectors only occasionally by small amounts.
    }
    \label{img:chains}
\end{figure}
We train \textit{R-iterative} with $N_R=3$ iterations of R and G per batch (resulting in four images per input vector) for 100k batches at a learning rate of $0.0005$, followed by a further 30k batches at a learning rate of $0.0001$. (This is a rather short time to train \textit{R-iterative}, which was chosen due to high training times and hardware constraints. We expect the model to be rather far away from convergence at that point.) $\lambda_R$ is set to $0.9$. Figure~\ref{img:r-iterative-chains} shows example results of the experiment. The model achieves continuous improvements with each iteration, which are similar to what one would expect to see in recurrent GAN architectures~\cite{RecurrentGAN}. This shows that merely changing the input vector $\mathbf{z}$ and feeding it back into the same generator can lead to improved image quality, similar to what other models achieve by directly altering the generated images.

\subsection{G-LIS}
\label{sec:experiments-G-LIS}
\begin{figure}[tp]
    \begin{subfigure}{1\textwidth}
        \centering
        \includegraphics[width=1\textwidth]{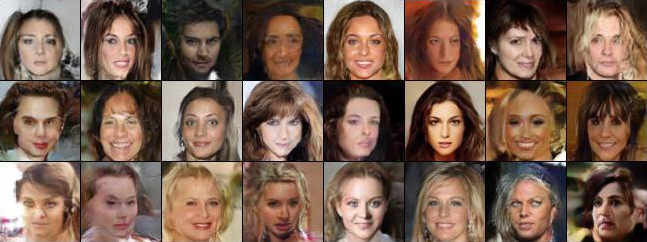}
        \caption{0 LIS modules}
        \label{img:glis-0-modules}
    \end{subfigure}
    \begin{subfigure}{1\textwidth}
        \centering
        \includegraphics[width=1\textwidth]{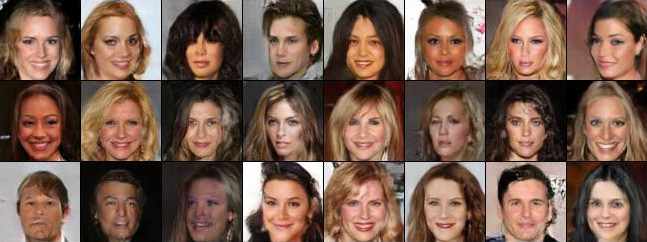}
        \caption{1 LIS module}
        \label{img:glis-1-modules}
    \end{subfigure}
    \begin{subfigure}{1\textwidth}
        \centering
        \includegraphics[width=1\textwidth]{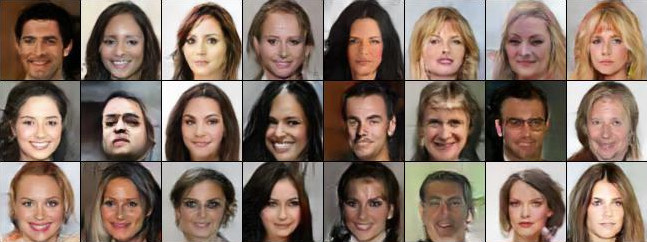}
        \caption{3 LIS modules}
        \label{img:glis-3-modules}
    \end{subfigure}
    \caption{
        Results of G-LIS with different numbers of LIS modules. All three experiments ran for 300k batches each.
        (a) Contains no LIS modules (baseline model). This is similar to a standard DCGAN-like~\protect\cite{DCGAN} generator with four convolutional layers (starting at 256 filters) and weight normalization~\protect\cite{WN}.
        (b) The exactly same generator as in (a) with one LIS module added to it.
        The frequency of large scale visual artifacts decreases significantly. (See Appendix~\protect\ref{sec:appendix-g-lis} for larger images, where the effect is more obvious.)
        As a negative side effect, the diversity of the backgrounds also decreases. We additionally see a stronger tendency towards frontalized faces.
        (c) The exactly same generator as in (a) with three LIS module added to it.
        Again, the frequency of large scale visual artifacts decreases significantly, but backgrounds become more uniform.
    }
    \label{img:glis-n-modules}
\end{figure}
We train \textit{G-LIS} with $\{0, 1, 3\}$ LIS modules, where $0$ is the baseline without error avoidance. $\lambda_R$ is set to $0.9$. Note that due to the exponentiation of $\lambda_R$ (the strength of the similarity constraint) per LIS module, the experiment with $1$ module has less room to change $\mathbf{z}$ than the one with $3$ modules. We train each experiment for 100k batches at a learning rate of $0.0005$, followed by 200k batches at a learning rate of $0.0001$. In contrast to \textit{R-iterative}, the generator is always only trained once per batch, making the results more comparable. Figure~\ref{img:glis-n-modules} shows example images. If no LIS module is used, large scale artifacts and completely broken images are still common after 300k batches. Adding one module significantly reduces the frequency of such artifacts, thereby increasing image quality. Even in large amounts of generated images we found virtually zero completely broken faces (see Appendix~\ref{sec:appendix-g-lis} for examples). Adding a total of three modules seemed to lead to a further -- but small -- improvement.%
%Interestingly, the improvement with $3$ modules is not as continuous (per module) as in the case of R-iterative. Instead, the second module has usually almost no effect, while
We did not observe an obvious decrease in diversity with regards to facial features. However, the diversity of backgrounds seemed to be significantly affected. When using three LIS modules, most backgrounds became uniform colors or simple color gradients. This effect is also clearly visible when using one module. We note that this is reminiscent of the outputs generated by BEGAN~\cite{BEGAN}. We further noticed that especially when using three LIS modules, the third one had a tendency to convert side-views of faces to frontal views. As non-frontal faces are uncommon in CelebA, it is expected that the generator is rather bad at creating them and that the discriminator will be more likely to view them as fakes (even at perfect quality, due to their lower prior probability). Hence, it is logical for the LIS modules to often frontalize them.

In another experiment, we test whether \textit{G-LIS} and \textit{R-separate} can be combined. We train \textit{R-separate} on the results of one of the previously developed G-LIS models (specifically, the one with a single LIS module). By applying R, we see a slight increase of D's loss, indicating higher image quality. Visually however, we did not perceive a clear effect. Positive and negative changes seemed to outweigh each other, making the application of \textit{R-separate} in this case not beneficial (see Appendix~\ref{sec:appendix-r-separate} for examples). It is possible that \textit{R-separate} is only useful when the generated images still contain a significant amount of artifacts. An alternative explanation would be that all achievable benefits of \textit{R-separate} were already gained by the LIS module.

By default, each LIS module uses a mean squared error to enforce similarity of $\mathbf{z}'$ with $\mathbf{z}$. To verify whether this is necessary, we train a model with three LIS modules and a $\lambda_R$ of $0$, thereby effectively deactivating the similarity constraint. During the first 100k batches at a learning rate of $0.0005$ we observe significant mode collapse that does not disappear throughout the training (see Appendix~\ref{sec:G-LIS} for an example image). We therefore consider the use of a similarity constraint to be essential, though its optimal strength has to be determined by further research.

One question that remains is whether the noise vectors generated by the LIS modules ($\mathbf{z}'$) really stay close to their original locations in the input space ($\mathbf{z}$), i.~e. whether the similarity constraint holds.
%One indication for this are the measured loss values of the similarity constraints, which usually stay below $\sim0.1$ for the first LIS modules and below $\sim0.15$ for the last module, which always had by far the highest loss (
To investigate this, we approximate and visualize the probability distributions of each component of $\mathbf{z}$, before and after applying LIS modules. (Appendix~\ref{sec:appendix-densities} contains six example plots.) The results indicate that the components remain normally distributed, with no added "bumps" or "valleys". Only the mean and standard deviations change. This however does not exclude that under specific conditions single vectors are moved. To analyze whether this is the case, we generate and visualize t-SNE~\cite{t-SNE} embeddings of the vectors before and after applying the LIS modules. (See Appendix~\ref{sec:appendix-embeddings} for plots.) We observe that the distributions of vectors seems to largely remain stable throughout the modules. Nevertheless, with each module, areas with higher and lower density become more prominent, especially at the last step. This supports the initial assumption that the implemented technique mostly keeps the input space and only moves vectors when it really has to, i.~e. in order to avoid artifacts.

To verify that our method is not only limited to CelebA, we run experiments on 102~flowers~\cite{102flowers}, 10k~cats~\cite{10kcats}, LSUN~churches~\cite{LSUN} and CIFAR-10~\cite{CIFAR10}. The resulting images show no signs of mode collapse or lack of diversity. Our best measured Inception Score~\cite{ImprovedGAN} on CIFAR-10 is 5.13. See Appendix~\ref{sec:appendix-g-lis} for example images and Appendix~\ref{sec:appendix-inception-scores} for inception scores. Subjectively, we got the best results on CelebA. That is expected, as generating human faces is a rather easy task and the dataset is large. In general, our method has two requirements to be expected to work well. First, the generator must produce at least some good-looking images. That is because the method does not correct errors, it avoids them by moving noise vectors from bad to good locations in the input space. If there are no good locations (i.~e. all generated images look bad), then error avoidance will not help. Second, the discriminator must have a decent understanding of what a good image looks like and how bad images can be improved. Otherwise the LIS modules will receive uninformative gradients and move noise vectors from bad locations to other bad locations.

\section{Related Work}
\label{sec:related-work}

Projecting backwards from images to noise vectors is not a new concept in GANs. Usually however, it is done to acquire noise vectors for real images. In these cases, training a reverser network (from fake images to noise vectors) is not enough, as the images created by the generator are usually be biased. Most authors instead choose to approximate input vectors on an image-by-image basis using gradient descent. In these cases, the desired output image and the generator's parameters are fixed. A distance between the currently generated image and the desired output is measured and backpropagated up to $\mathbf{z}$, which is changed according to the gradient. This technique is employed in \cite{WN, GenerativeVisualManipulation, PreciseRecovery, InvertingTheGenerator}. In \cite{PreciseRecovery} the authors notice that the reverse projection can be used to remove gaussian noise, but do not investigate any potential for artifact correction. In \cite{GenerativeVisualManipulation} a reverser is explicitly added, similar to the one in this paper, but only used to speed up the prediction of $\mathbf{z}'$.

Iterative architectures, like the one in \textit{R-iterative}, are occasionally~\cite{StackGAN, StackedGAN, RecurrentGAN} described in the literature. GRAN~\cite{RecurrentGAN} works recurrently and generates residual images over multiple timesteps that are summed elementwise in order to derive the final image. It has significant similarity with \textit{R-iterative} as it uses at each timestep one network to project the previous results onto a vector (similar to R) and a second network to convert the vector to an image (similar to G). In contrast to \textit{R-iterative}, the generated vectors are not constrained to be close to $\mathbf{z}$, though the use of residual outputs might have an indirect constraining effect. StackGAN~\cite{StackGAN} is another iterative architecture. It uses two timesteps, each of them consisting of its own generator and discriminator. The timesteps work at different scales, making the architecture roughly comparable to an end-to-end trained LAPGAN~\cite{LAPGAN}. Information is transferred between the timesteps by using several convolutions to transform the first image into a high-dimensional low-resolution space (of shape $\mathbb{R}^{512\times16\times16}$), which is effectively $\mathbf{z}'$ for the second timestep. In contrast to \textit{R-iterative}, no explicit similarity constraint is used. However, $\mathbf{z}'$ is generated using convolutions, which are translation-invariant and transform only small local neighbourhoods, making them naturally constrained. As such, we expect the convolutions between the two stages to work simililarly to R in \textit{R-iterative}, leading to an error avoiding effect by choosing a good input space for the second stage. Due to the use of convolutions, that error avoidance will likely work predominantly on a local level, making it harder to avoid global image errors. This is also observed by the authors of StackGAN, who noted that the main source of errors in the second stage are global errors of the first stage (broken rough shapes or bad choices of image colors). We expect some of these errors to be fixed by adding an LIS module to the first stage in order to prevent such global errors early on. We further speculate that the main benefit of the second stage might be error avoidance by choosing good input spaces. If that would be the case, the architecture could be collapsed to a single generator and discriminator, saving computation time and memory.

\section{Conclusion}

In this paper we suggested to manipulate the input space of generative adversarial networks in order to enable them to avoid making errors. We derived three possible models from that idea. \textit{R-separate} worked by applying a simple regression from $\mathbf{x}_{fake}$ to $\mathbf{z}$. This was extended in \textit{R-iterative} by also optimizing on D and enabling multiple iterations. The architecture was then collapsed into G-LIS using LIS modules, which consisted of residual fully connected layers with similarity constraints. All three architectures showed improvements in image quality. These were most prominent in the case of G-LIS, which managed to significantly reduce the frequency of large scale artifacts, while the loss in diversity remained mostly limited to backgrounds. As LIS modules are nearly free with respect to computation and memory, we consider them a useful addition to almost any GAN.

\bibliographystyle{plain}
\bibliography{paper}

% -------------------------------

\clearpage
\begin{appendices}

\section{R-separate}
\label{sec:appendix-r-separate}

\begin{figure}[h]
    \centering
    \includegraphics[width=1\textwidth]{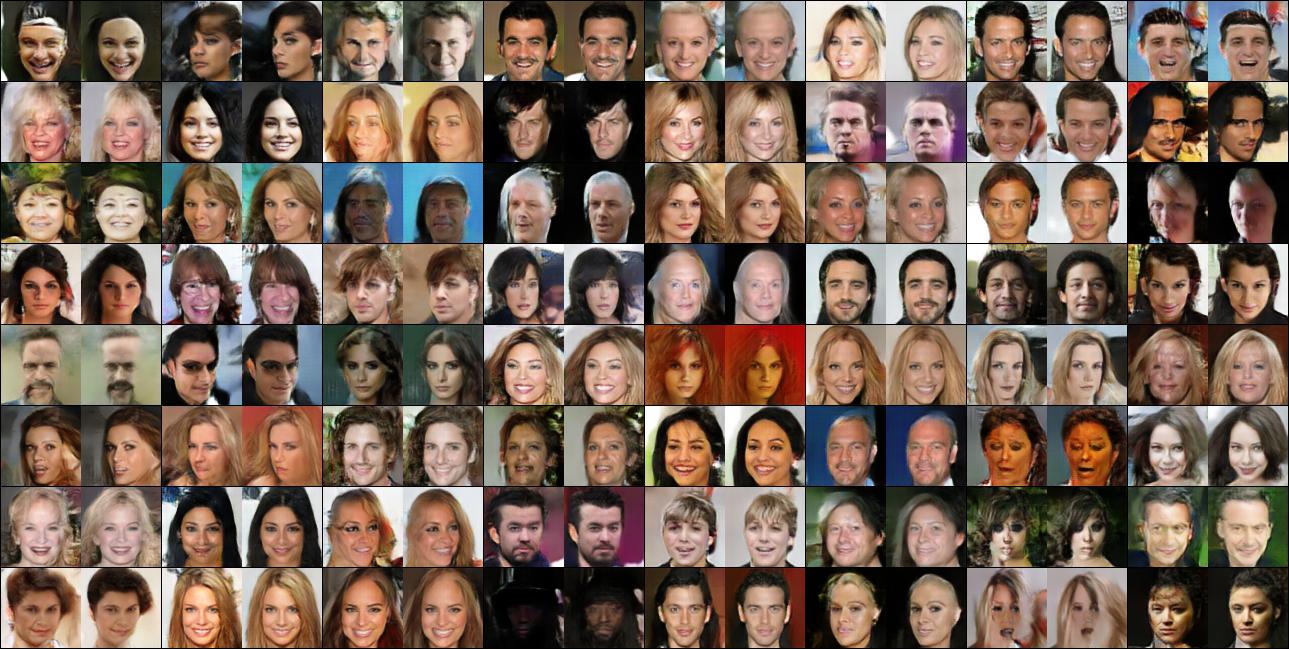}
    \caption{
        Example results of \textit{R-separate}. The generator (without any LIS modules) was trained for 50k batches (short training, which still results in many artifacts). The reverser was trained for 2.5k batches with spatial dropout of 10\% between convolutional layers. Applying \textit{R-separate} improves the images by repairing artifacts. This does not lead to blurriness and does not necessarily lead to the loss of small-scale features.
    }
    \label{img:appendix-rseparate-exp02}
\end{figure}

\begin{figure}[h]
    \centering
    \includegraphics[width=1\textwidth]{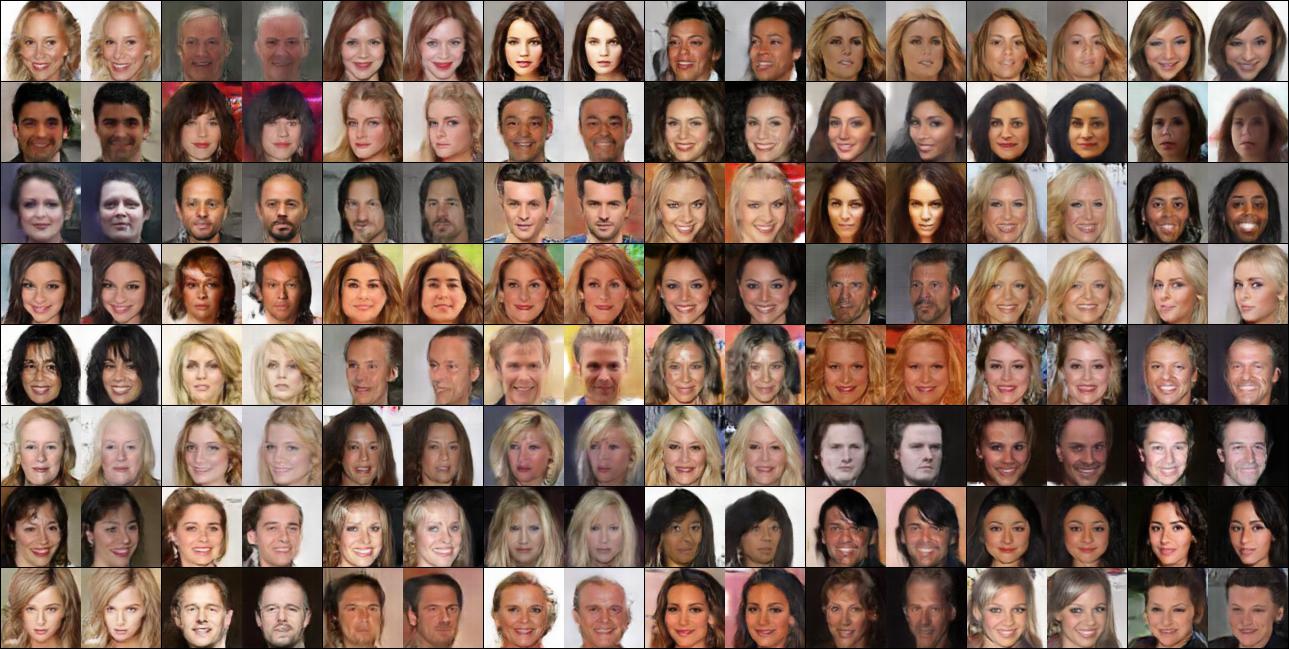}
    \caption{
        \textit{R-separate} applied to G-LIS with one LIS module, after being trained for 300k batches. \textit{R-separate} was trained for 2.5k batches without spatial dropout. We see occasionally minor improvements. It is not clear whether the positive effects outweigh the negative ones.
    }
    \label{img:appendix-rseparate-exp04}
\end{figure}

% ---------------------------------------

\clearpage
\section{R-iterative}
\label{sec:appendix-r-iterative}

\begin{figure}[h]
    \centering
    \includegraphics[width=1\textwidth]{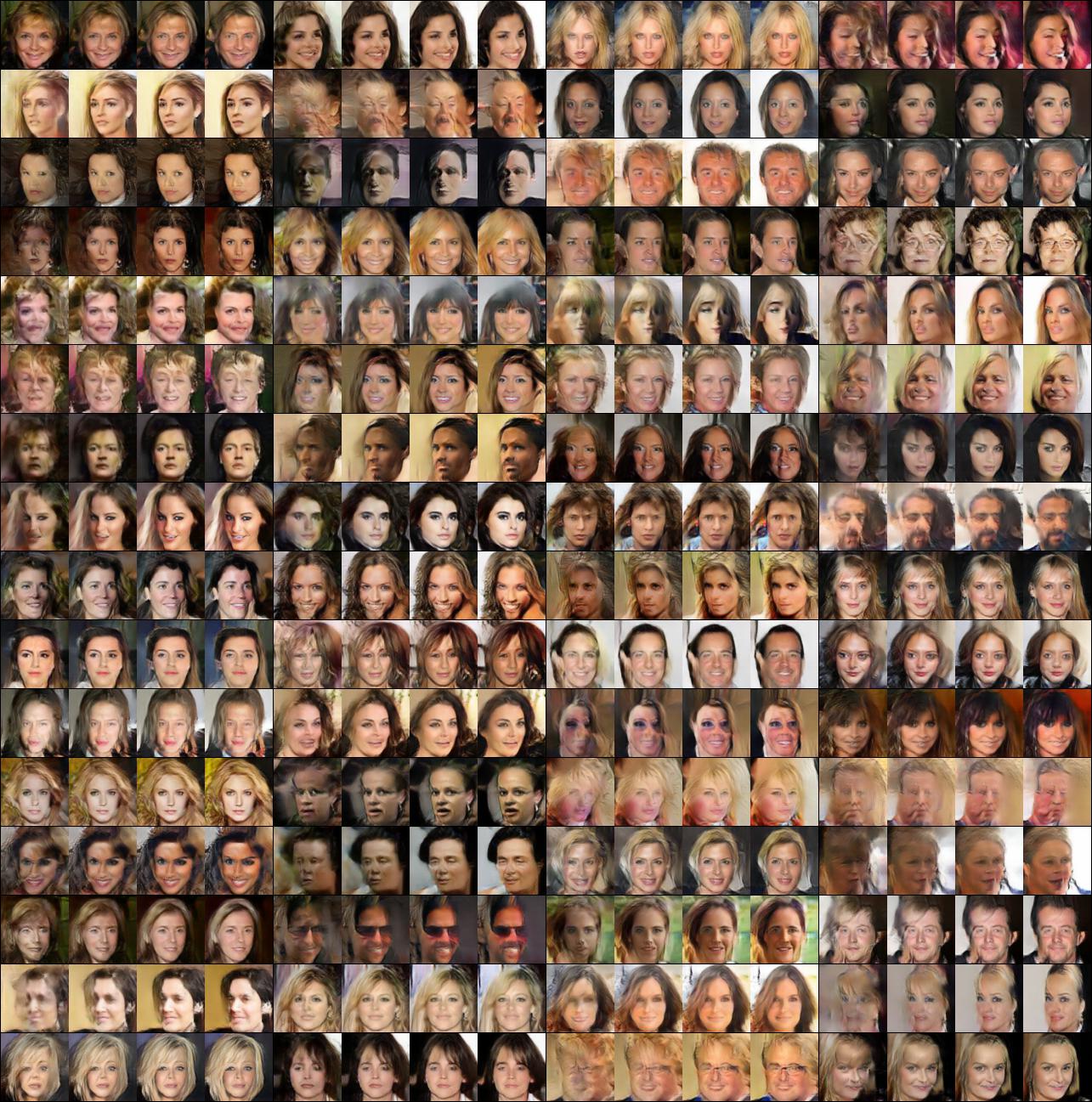}
    \caption[R-Iterative at three iterations (large)]{
        Example results of \textit{R-iterative}. The model was trained according to the descriptions in section~\protect\ref{sec:experiments-R-iterative}. Each image per group of images shows the result of one iteration of the model. At each iteration, the exactly same generator is applied (i.~e. same parameters) and merely the input vector is changed. That means, that one could instantly generate the fourth iteration's image, if the respective input vector was known. It also means that each column resembles the results from one unique input space (though all input spaces are very similar). Note that the final images are of rather low quality as the training time was kept short (130k batches) due to high computational costs.
    }
    \label{img:appendix-riterative-chains}
\end{figure}

\begin{figure}[h]
    \includegraphics[width=1\textwidth]{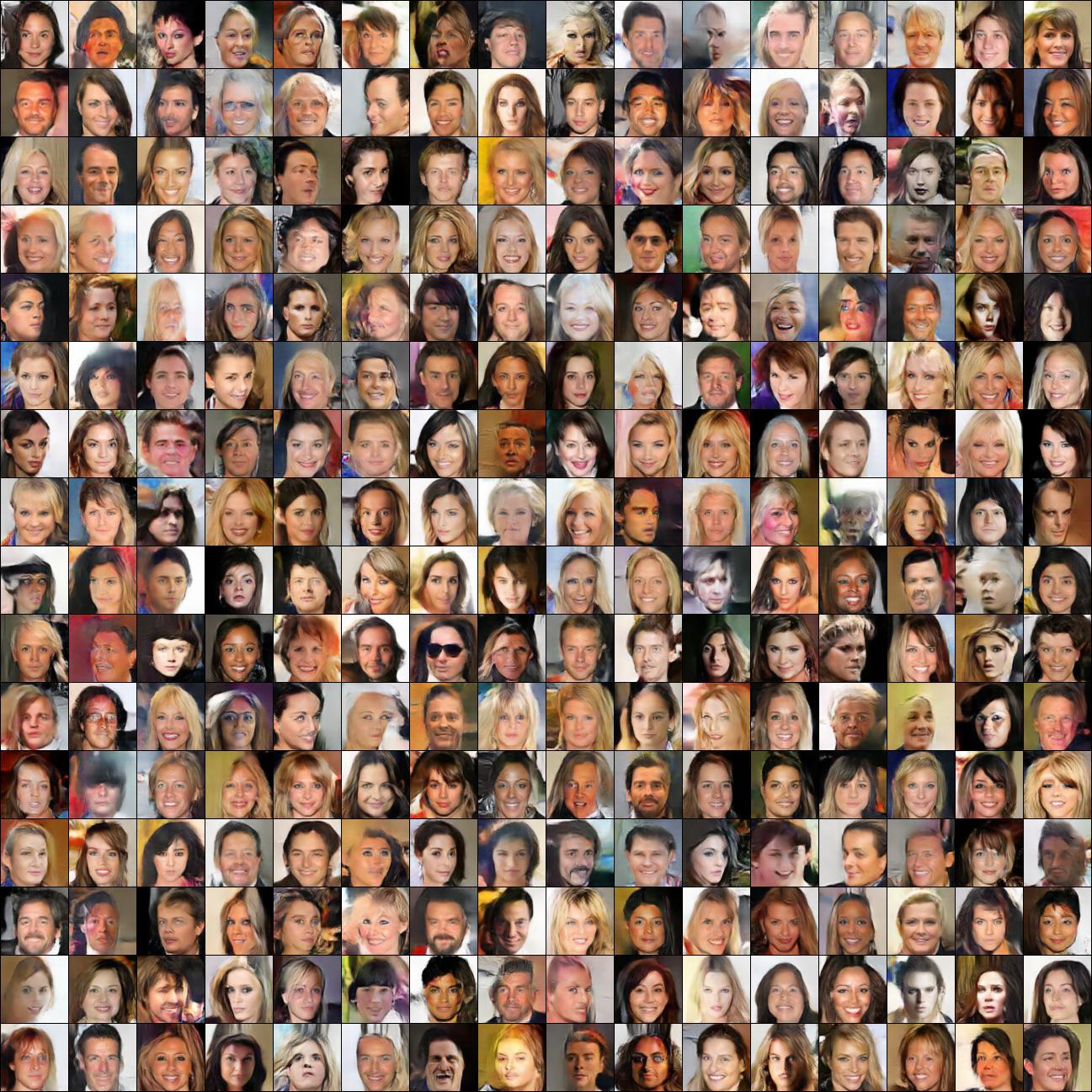}
    \caption{
        Example results of \textit{R-iterative} using the same model as before. Shown are only the results of the final iteration. As before, note that the training time of the model was kept short (130k batches).
    }
    \label{img:appendix-riterative-last}
\end{figure}

% -----------------------------------------

\clearpage
\section{G-LIS}
\label{sec:appendix-g-lis}

\begin{figure}[h]
    \centering
    \includegraphics[width=1\textwidth]{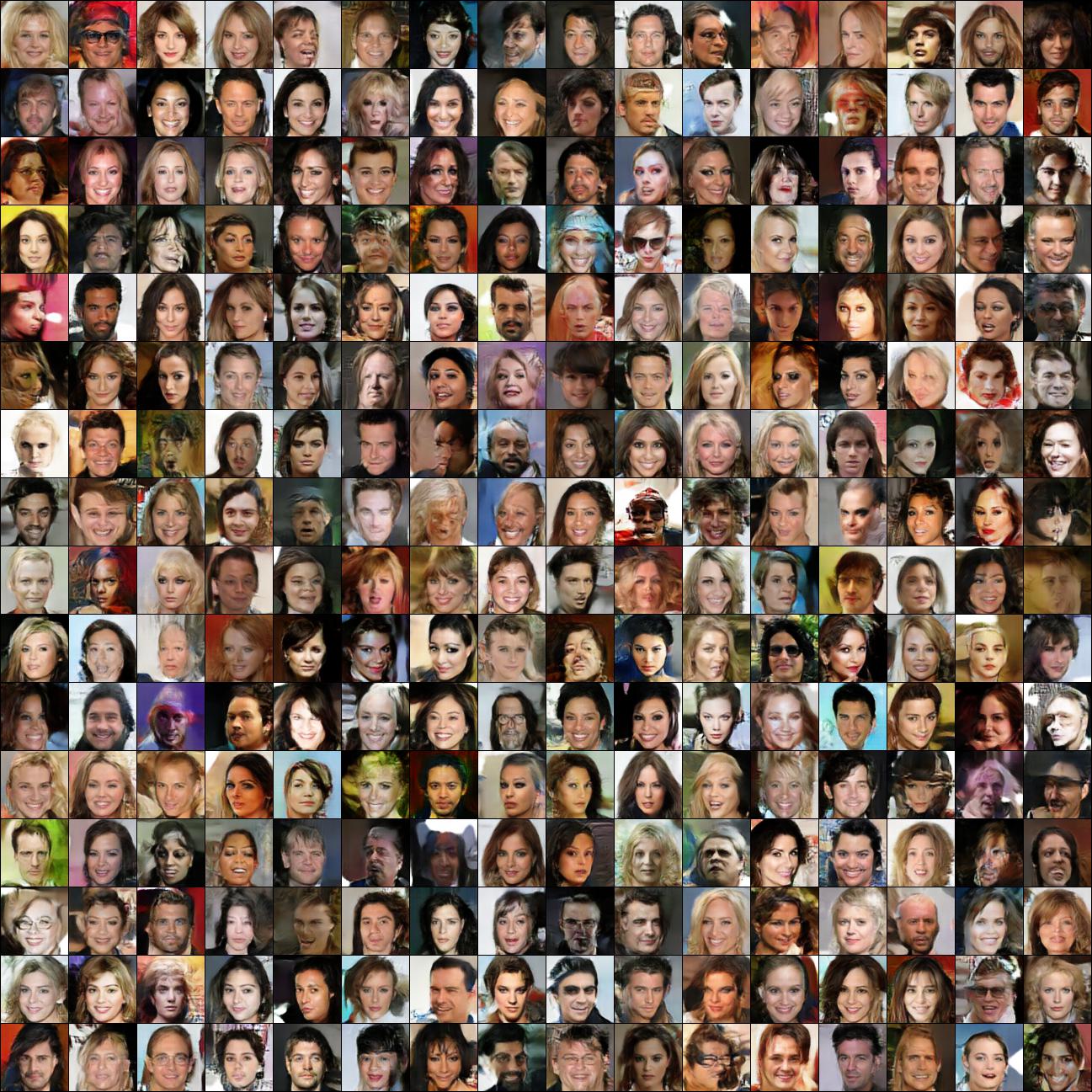}
    \caption[G-LIS with 0 LIS modules (large)]{
        Results of \textit{G-LIS} with zero LIS modules. Training happened according to the descriptions in section~\protect\ref{sec:experiments-G-LIS}. This is similar to the models presented in~\protect\cite{WN}, though here at $80\times80$ resolution and with one less layer in G and D. The results are used as a baseline to compare against (see the following images). When zooming in, one can see that many images have large scale artifacts, sometimes to the point where the face is no longer recognizable. (We note that the results in~\protect\cite{WN} contain less artifacts, which we assume comes from higher number of layers and parameters. The difference in quality is here rather unimportant, as the experiments are primarily supposed to show the effects of using error avoidance.)
    }
    \label{img:appendix-glis-riter0-final}
\end{figure}

\begin{figure}[h]
    \centering
    \includegraphics[width=1\textwidth]{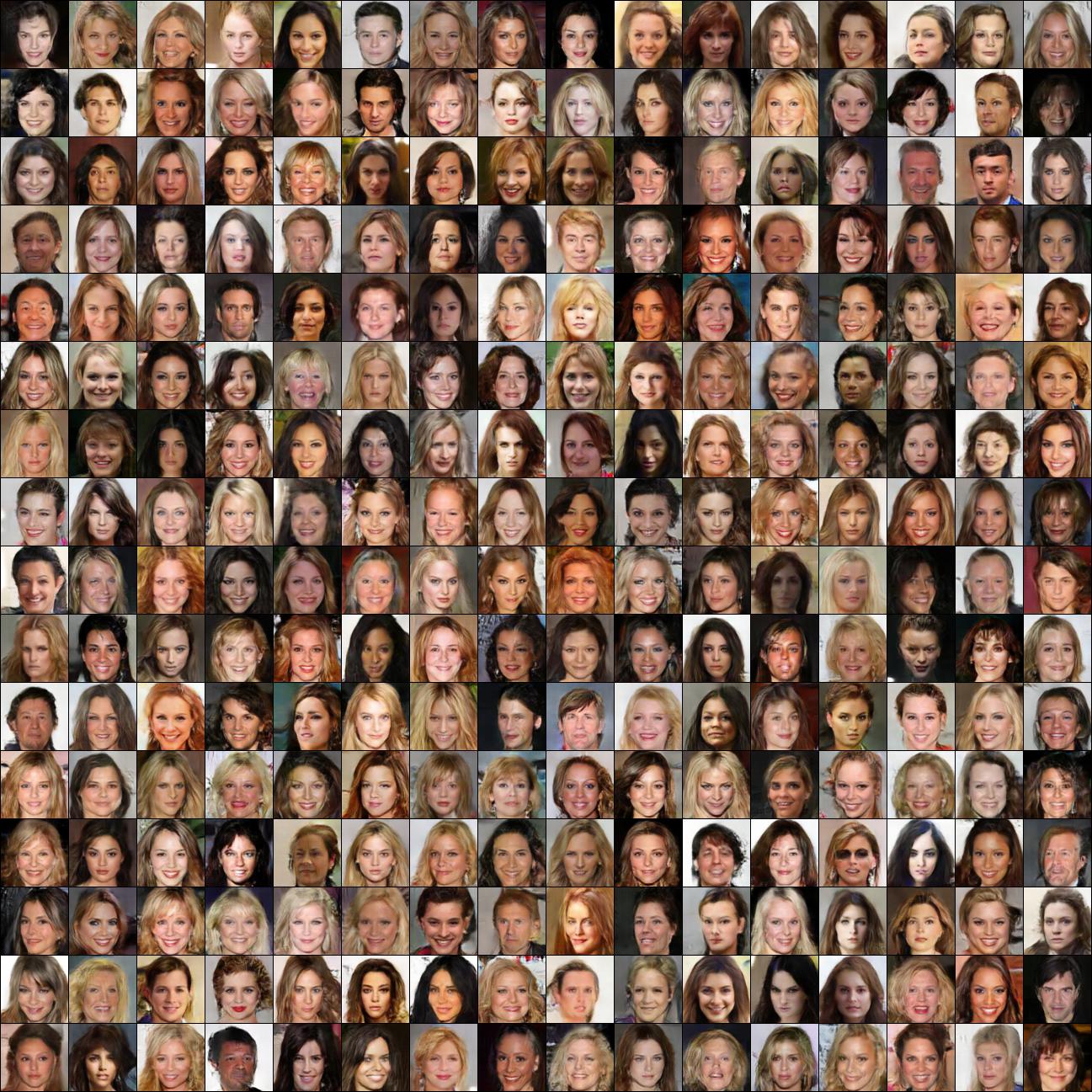}
    \caption[G-LIS with 1 LIS module (large)]{
        Results of \textit{G-LIS} with one LIS module. Training happened according to the descriptions in section~\protect\ref{sec:experiments-G-LIS}. Note that there is not a single completely broken face (results are not cherry-picked) -- even though the core of the generator was not changed and the LIS module does nothing more than picking better input vectors.
    }
    \label{img:appendix-glis-riter1-final}
\end{figure}

\begin{figure}[h]
    \centering
    \includegraphics[width=1\textwidth]{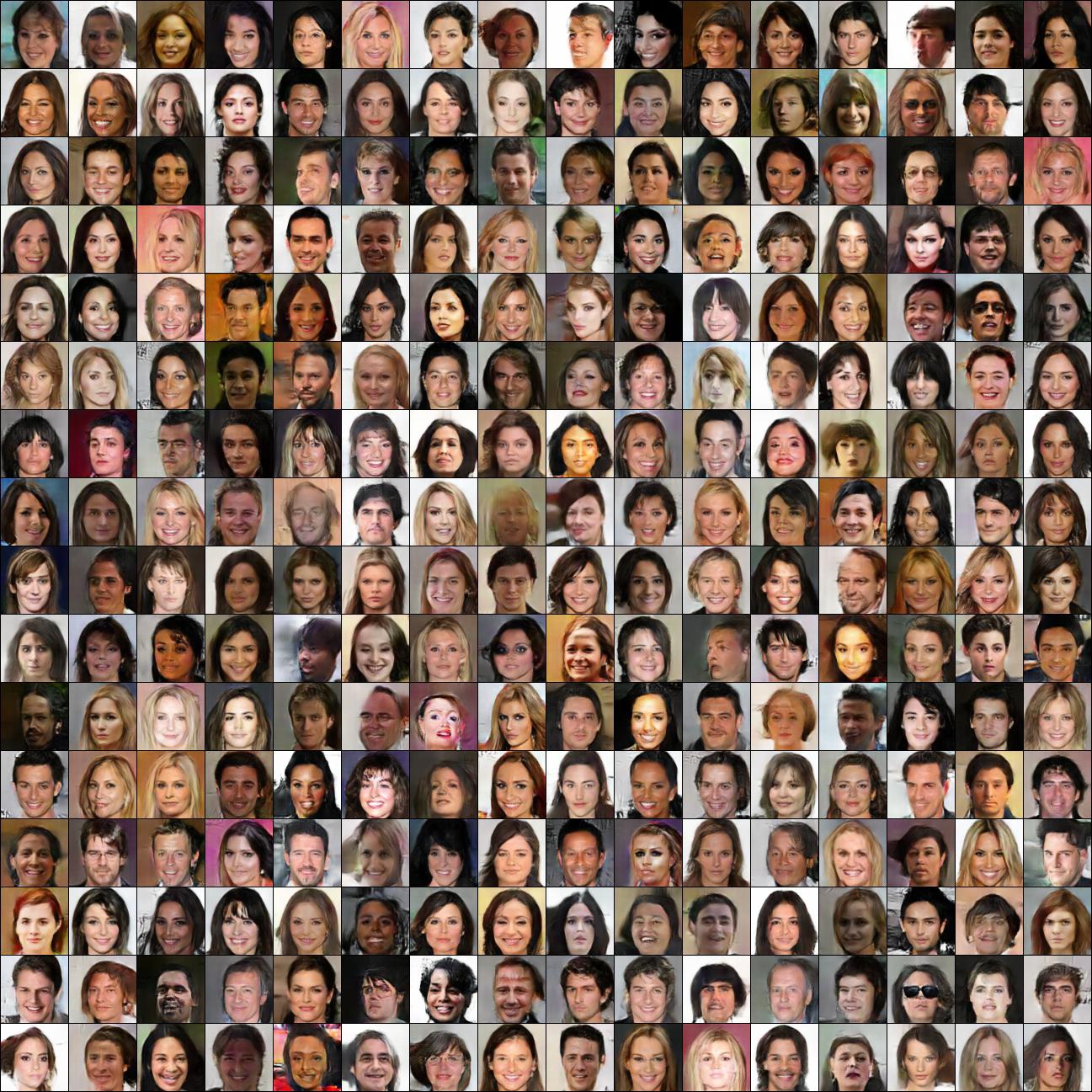}
    \caption[G-LIS with 3 LIS modules (large)]{
        Results of \textit{G-LIS} with three LIS modules. Training happened according to the descriptions in section~\protect\ref{sec:experiments-G-LIS}. Note the more uniform backgrounds than in the case of \textit{G-LIS} with zero or one LIS modules. We also noticed more frontalized faces (less sideviews) and a stronger bias towards dark hair.
    }
    \label{img:appendix-glis-riter3-final}
\end{figure}

\begin{figure}[h]
    \centering
    \includegraphics[width=1\textwidth]{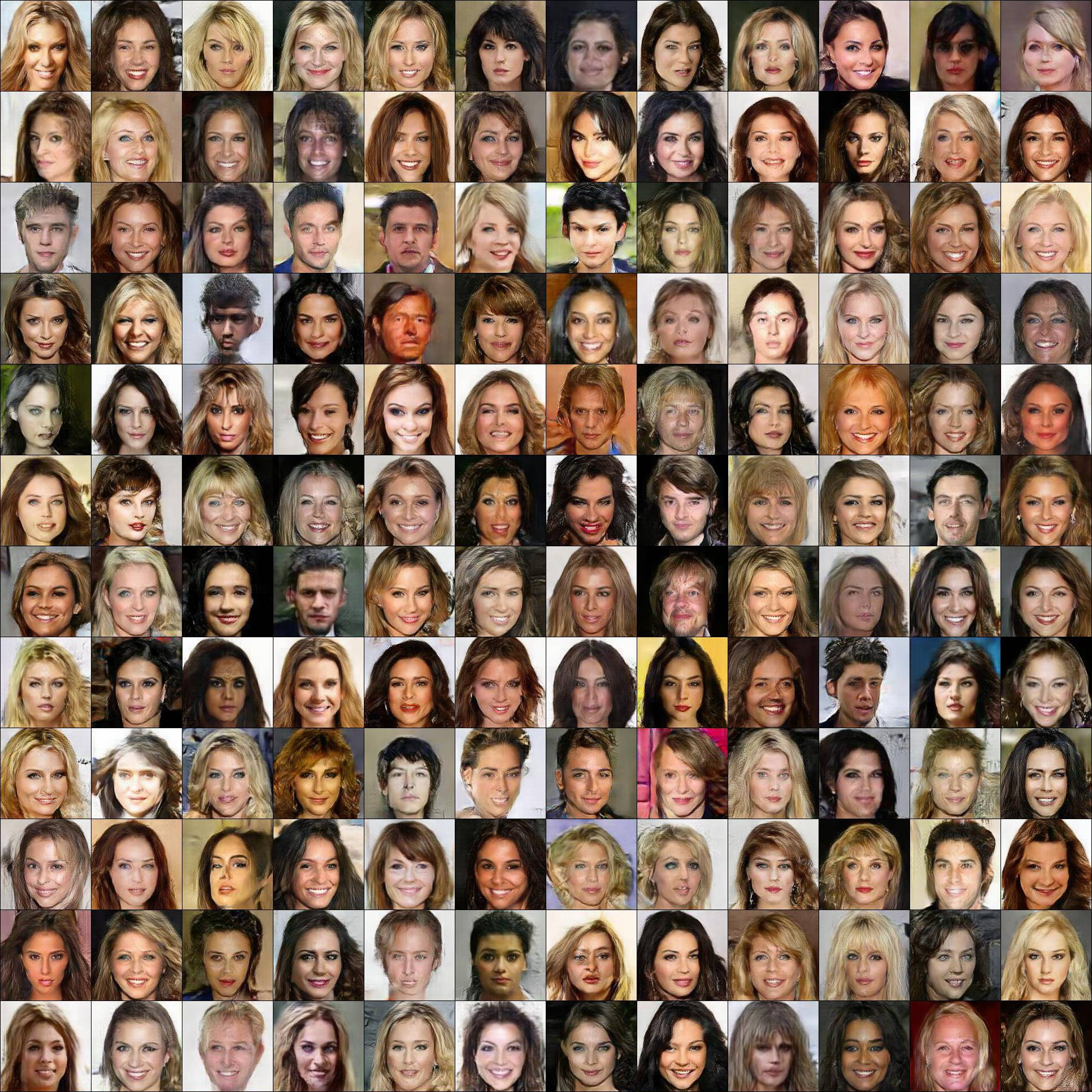}
    \caption{
        Results of \textit{G-LIS} with one LIS module, trained on CelebA at $160\times160$ resolution for 400k batches at a learning rate of~$0.00002$, followed by 170k batches at a learning rate of~$0.00001$.
        This seemed to result in slightly more large scale artifacts than the corresponding network at a resolution of $80\times80$.
    }
    \label{img:appendix-glis-riter1-160x160-final}
\end{figure}

\begin{figure}[h]
    \centering
    \includegraphics[width=1\textwidth]{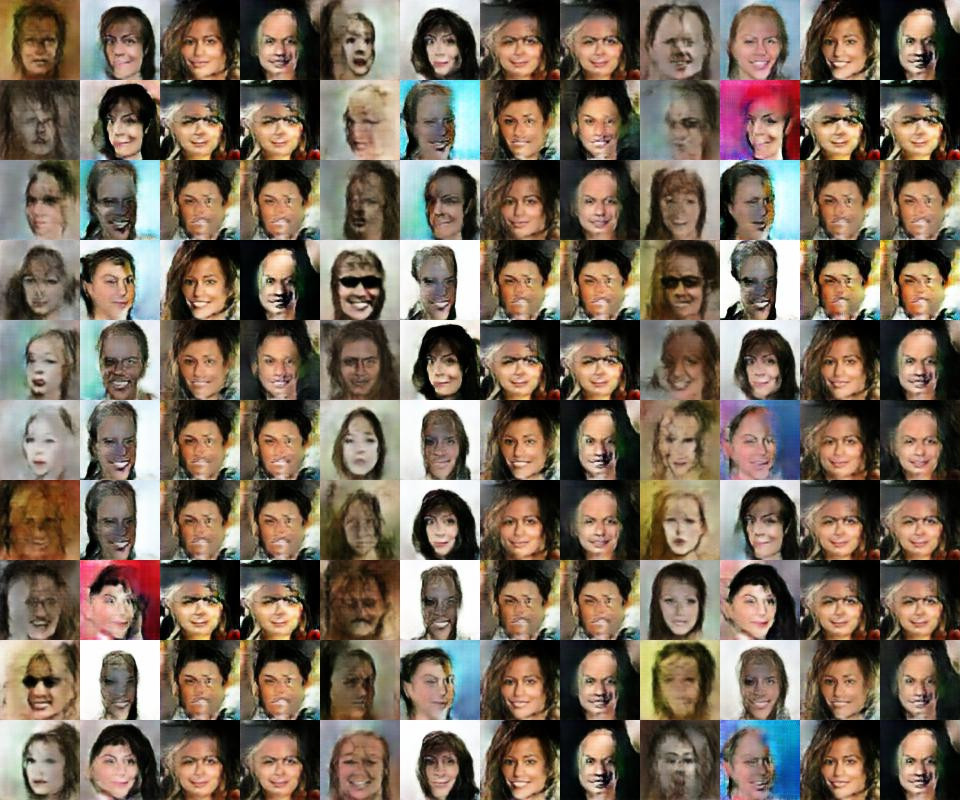}
    \caption{
        Results of \textit{G-LIS} with three LIS modules, trained on CelebA for 100k batches at a learning rate of~$0.00005$. Images for the original noise vector and all LIS modules are shown (i.~e. four images per noise vector). The strength of $\lambda_R$ was set to $0$ in this experiment, which means that the similarity constraint on the LIS modules was deactivated. This caused the generator to show significant mode collapse throughout the whole training, indicating that the constraint is essential for the method to work.
    }
    \label{img:appendix-glis-riter3-mode-collapse}
\end{figure}

\clearpage

\begin{figure}[h]
    \begin{subfigure}{1\textwidth}
        \centering
        \includegraphics[width=1\textwidth]{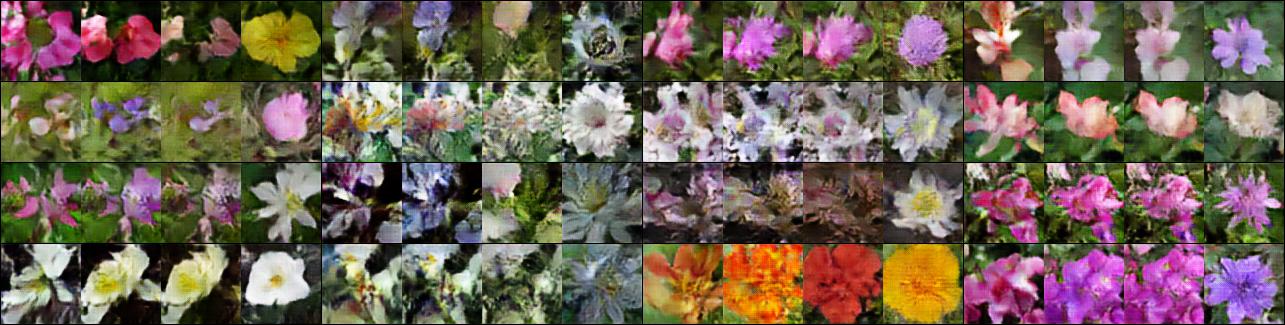}
        \caption{
            Original noise vector and LIS modules 1 to 3
        }
        \label{img:appendix-glis-riter3-120flowers-chains}
    \end{subfigure}
    \begin{subfigure}{1\textwidth}
        \centering
        \includegraphics[width=0.75\textwidth]{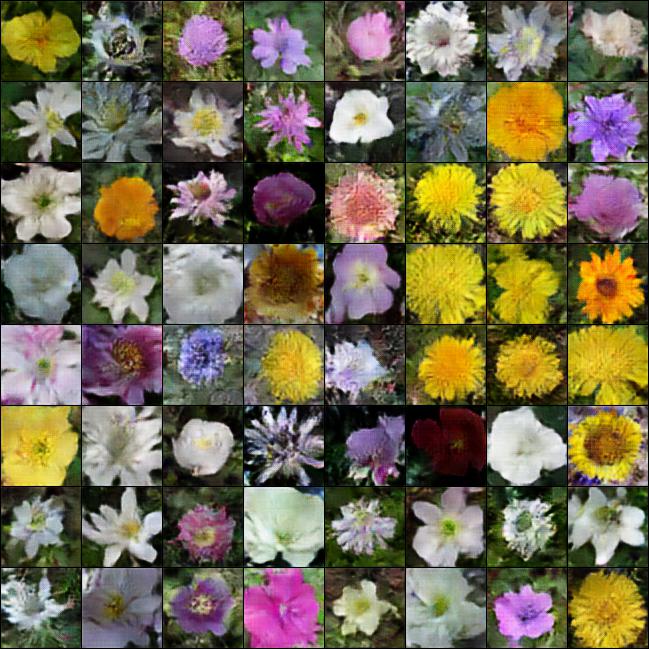}
        \caption{LIS module 3}
        \label{img:appendix-glis-riter3-120flowers-final}
    \end{subfigure}
    \caption[G-LIS with 3 LIS modules (102 flowers, large)]{
        Results of \textit{G-LIS} with three LIS modules, trained on 102~Flowers~\protect\cite{102flowers} at $80\times80$ resolution for 500k batches at a learning rate of~$0.00002$, followed by 100k batches at a learning rate of~$0.00001$.
        (a) Shows the images generated when using the original noise vector $\mathbf{z}$ as well as the vectors produced by each LIS module $\mathbf{z}_i'$, leading to four images per noise vector.
        (b) Shows the results when using only the final LIS module's noise vectors.
    }
    \label{img:appendix-glis-riter3-120flowers}
\end{figure}

\clearpage

\begin{figure}[h]
    \begin{subfigure}{1\textwidth}
        \centering
        \includegraphics[width=1\textwidth]{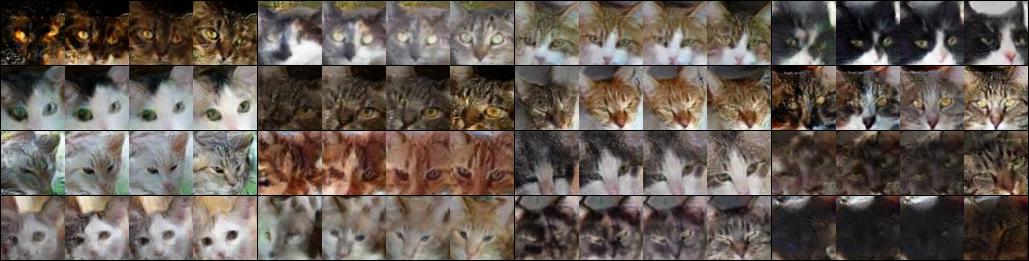}
        \caption{
            Original noise vector and LIS modules 1 to 3
        }
        \label{img:appendix-glis-riter3-10kcats-chains}
    \end{subfigure}
    \begin{subfigure}{1\textwidth}
        \centering
        \includegraphics[width=0.75\textwidth]{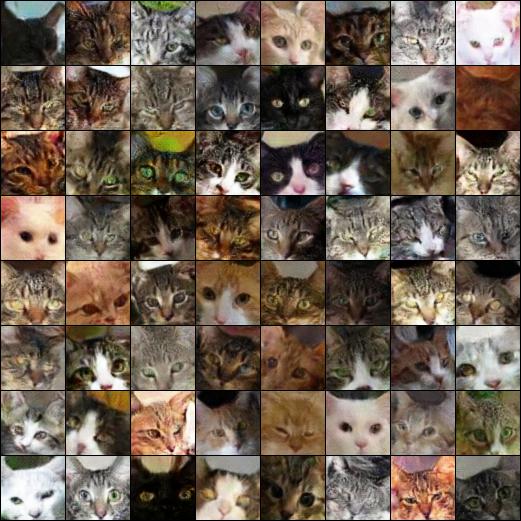}
        \caption{LIS module 3}
        \label{img:appendix-glis-riter3-10kcats-final}
    \end{subfigure}
    \caption[G-LIS with 3 LIS modules (10k cats, laarge)]{
        Results of \textit{G-LIS} with three LIS modules, trained on 10k~Cats~\protect\cite{10kcats} at $64\times64$ resolution for 700k batches at a learning rate of~$0.00002$.
        (a) Shows the images generated when using the original noise vector $\mathbf{z}$ as well as the vectors produced by each LIS module $\mathbf{z}_i'$, leading to four images per noise vector.
        (b) Shows the results when using only the final LIS module's noise vectors.
    }
    \label{img:appendix-glis-riter3-10kcats}
\end{figure}

\clearpage

%\begin{figure}[h]
%    \includegraphics[width=1\textwidth]{images/glis-exp13-churches-riter1-chains.jpg}
%    \caption{
%        Results of \textit{G-LIS} with one LIS module, trained on LSUN~churches~outdoor~\cite{LSUN} at $80\times80$ resolution for 500k batches at learning rate $0.00002$.
%    }
%    \label{img:appendix-glis-riter1-churches-chains}
%\end{figure}

\begin{figure}[h]
    \centering
    \includegraphics[width=1\textwidth]{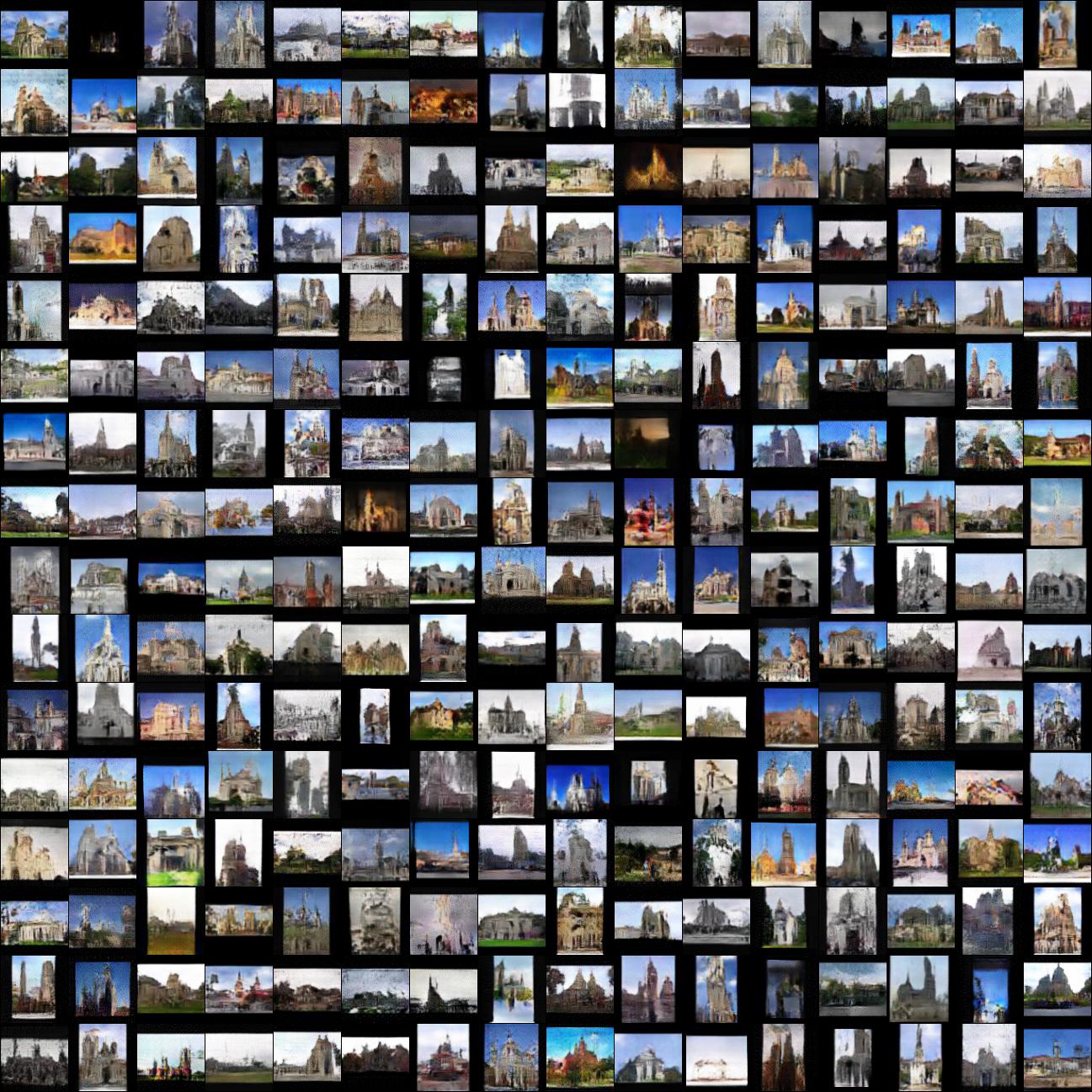}
    \caption[G-LIS with 1 LIS module (LSUN churches, large)]{
        Results of \textit{G-LIS} with one LIS module, trained on LSUN~churches~outdoor~\protect\cite{LSUN} at $80\times80$ resolution for 500k batches at a learning rate of~$0.00002$. Each shown image corresponds to the noise vector generated by the last LIS module. (For simplicity this experiment reused the $80\times80$ models from CelebA. Due to the often long and parallel edges of the churches, larger convolution kernels of e.~g. $5\times5$ would probably yield more realistic images in this case.)
    }
    \label{img:appendix-glis-riter1-churches-final}
\end{figure}

\clearpage

\begin{figure}[h]
\begin{tabular}{cc}
\subcaptionbox{After 0 LIS modules}{\includegraphics[width=0.45\textwidth]{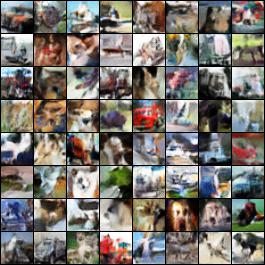}} &
\subcaptionbox{After 1 LIS modules}{\includegraphics[width=0.45\textwidth]{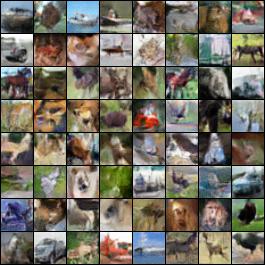}} \\
\subcaptionbox{After 2 LIS modules}{\includegraphics[width=0.45\textwidth]{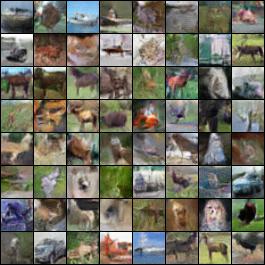}} &
\subcaptionbox{After 3 LIS modules}{\includegraphics[width=0.45\textwidth]{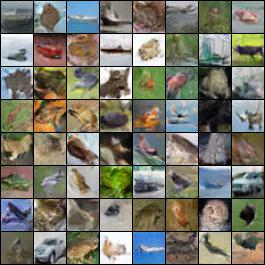}}
\end{tabular}
\caption[G-LIS with 3 LIS modules (CIFAR-10)]{
    Results of \textit{G-LIS} with three LIS modules, trained on CIFAR-10~\protect\cite{CIFAR10} at $32\times32$ resolution for one million batches at a learning rate of~$0.00002$. Each subimage shows the generated images when using the $i$-th LIS module's output $\mathbf{z}_i'$. The first image references the original noise vectors ($\mathbf{z}$). Using only two of the three LIS modules (third image) seemed to subjectively produce the best results, while the outputs of module three appeared to have the lowest amount of high frequency noise. The discriminator had the highest loss when using zero or three modules (first and last image). (Minor sidenote: The model used was very similar to the one in the $80\times80$ CelebA examples, hence not well optimized for $32\times32$ images.)
}
\label{img:appendix-glis-riter3-cifar10}
\end{figure}

% -------------------------------------

\clearpage
\section{Inception Scores}
\label{sec:appendix-inception-scores}

The following tables contain Inception Scores~\cite{ImprovedGAN} of G-LIS models trained on CelebA~\cite{CelebA} and CIFAR-10~\cite{CIFAR10}. Overall, we observed only a minor correlation between measured scores and perceived visual qualities. In the case of CelebA, severely broken images were rated as almost photo-realistic, while mode collapsed images achieved nearly the same score as the results of well-working models. We note that the Inception Score is based on the Inception model's last layer, which is supposed to be the most abstract one. Due to this, it is also expected to be robust towards noise and distortions, making the score invariant towards artifacts.

\begin{figure}[h]
\centering
\begin{tabular}{cc}
Model & Inception Score \\ \hline
Real Images ($160\times160$ center crops) & 3.29 \\ \hline
G-LIS 0 of 0 & 1.84 \\ \hline
G-LIS 0 of 1 & 2.13 \\
G-LIS 1 of 1 & 1.62 \\ \hline
G-LIS 0 of 3 & 2.58 \\
G-LIS 1 of 3 & 1.86 \\
G-LIS 2 of 3 & 1.77 \\
G-LIS 3 of 3 & 1.83 \\ \hline
G-LIS 3 of 3 with $\lambda_R=0$ & 1.51
\end{tabular}
\caption[Inception Scores on CelebA]{
    Inception Scores of four different G-LIS models trained on CelebA. All scores were estimated based on 50k images upscaled to $299\times299$. \textit{G-LIS M of N} references the images corresponding to the $M$-th of $N$ LIS modules, where $M=0$ is equivalent to the original noise vectors. \textit{G-LIS 0 of 0} is a baseline without LIS modules. The last model was trained without a similarity constraint ($\lambda_R=0$), leading to significant mode collapse. The training schedules are described in section~\protect\ref{sec:experiments}.
}
\label{tbl:appendix-celeba-inception-scores}
\end{figure}

\begin{figure}[h]
\centering
\begin{tabular}{cc}
Model & Inception Score \\ \hline
Real Images & 11.41 \\ \hline
Real Images ($28\times28$ center crops) & 9.70 \\ \hline
G-LIS 0 of 3 & 4.59 \\ \hline
G-LIS 1 of 3 & 5.13 \\
G-LIS 2 of 3 & 4.88 \\
G-LIS 3 of 3 & 5.02 \\
\end{tabular}
\caption{
    Inception Scores of a G-LIS model trained on CIFAR-10 for one million batches. The training dataset consisted of~$28\times28$ crops.
}
\label{tbl:appendix-cifar10-inception-scores}
\end{figure}

\begin{figure}[h]
    \centering
    \includegraphics[width=0.7\textwidth]{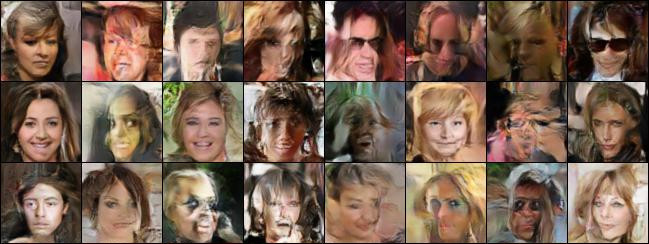}
    \caption[Examples of Inception Score failures]{
        Example images generated by G-LIS trained with three LIS modules, using directly the original noise vectors (i.~e. without executing any LIS modules). This matches the entry \textit{G-LIS 0 of 3} in table~\protect\ref{tbl:appendix-celeba-inception-scores}. The Inception Score of these results is $2.58$, close to real images with $3.2$. The true visual quality however is low and most images are severely broken.
    }
    \label{img:appendix-inception-score-failure}
\end{figure}

% -------------------------------------

\clearpage
\section{Interpolations}
\label{sec:appendix-interpolations}

\begin{figure}[h]
\centering
\begin{tabular}{c}
\includegraphics[width=0.95\textwidth]{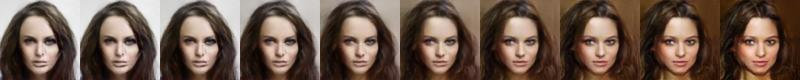} \\
\includegraphics[width=0.95\textwidth]{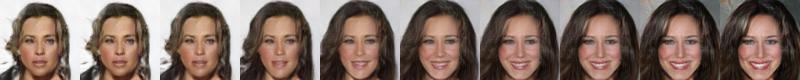} \\
\includegraphics[width=0.95\textwidth]{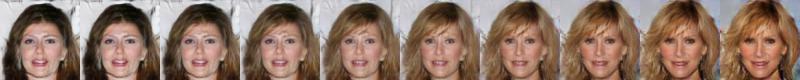} \\
\includegraphics[width=0.95\textwidth]{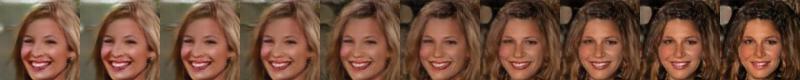} \\
\includegraphics[width=0.95\textwidth]{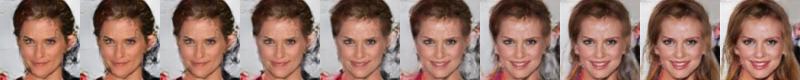} \\
\includegraphics[width=0.95\textwidth]{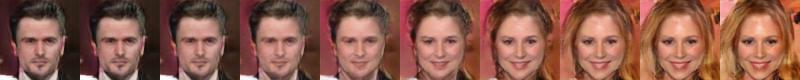} \\
\includegraphics[width=0.95\textwidth]{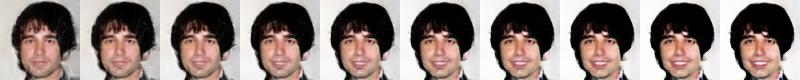} \\
\end{tabular}
\caption{
    Examples of linear interpolations. The images were generated by G-LIS with one LIS module, trained on CelebA for 300k batches (see experiments). Two noise vectors were randomly sampled as start and end points. Then, eight equally spaced noise vectors between these two were generated via linear interpolation. The ten noise vectors were fed into the generator to create the images. The results show that the model can handle even difficult transitions smoothly, e.~g. changes in camera angle (row one), zoom (row two), hair color (row three) or gender (row six). This indicates that the model did not overfit, despite being able to modify the input space.
}
\label{img:appendix-interpolations}
\end{figure}

% ----------------------------------------

\clearpage
\section{Perturbations}
\label{sec:appendix-perturbations}

\begin{figure}[h]
\centering
\begin{tabular}{c}
\includegraphics[width=0.55\textwidth]{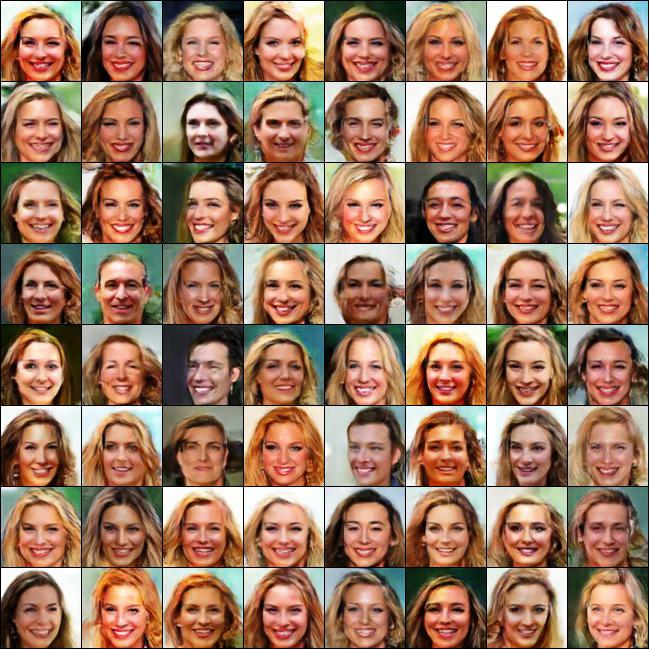} \\
\includegraphics[width=0.55\textwidth]{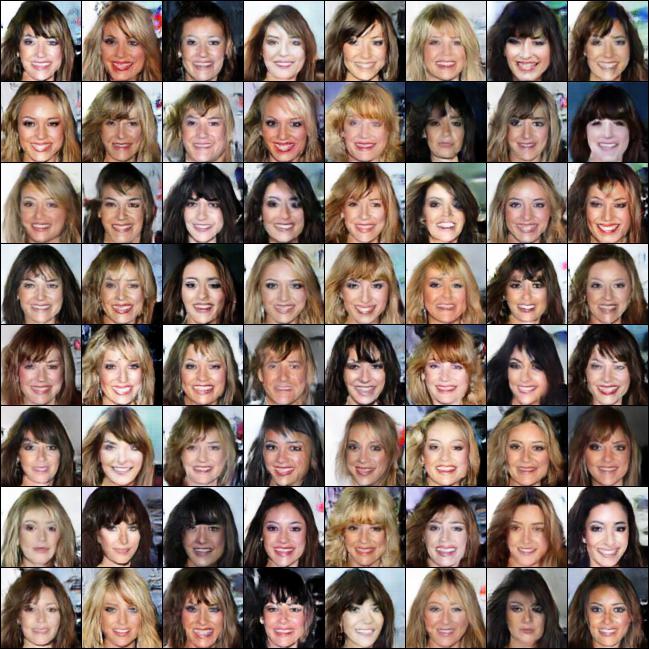}
\end{tabular}
\caption{
    Examples of perturbations. The images were generated by G-LIS with one LIS module, trained on CelebA for 300k batches (see experiments, same model as before for the interpolations). For each group of images, a random noise vector was sampled. For each subimage in these groups, samples from the standard normal distribution were added to each component of the initial noise vector. This led to 64 images with slightly different noise vectors per group. They were fed into the generator to create the images. The results indicate that the generator is able to create large numbers of plausible and different images with similar features. This indicates that the model did not overfit or simply memorize good-looking images.
}
\label{img:appendix-perturbations}
\end{figure}

% --------------------------------------

\clearpage
\section{2D embeddings of noise vectors}
\label{sec:appendix-embeddings}

\begin{figure}[h]
\begin{tabular}{cc}
\subcaptionbox{After 0 LIS modules}{\includegraphics[width=0.45\textwidth]{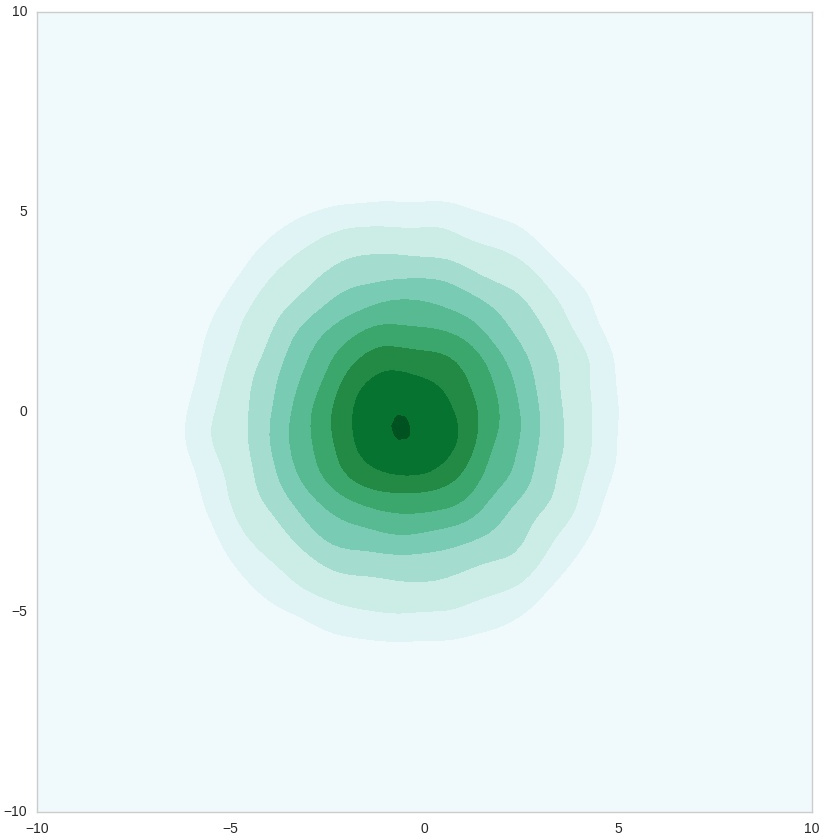}} &
\subcaptionbox{After 1 LIS modules}{\includegraphics[width=0.45\textwidth]{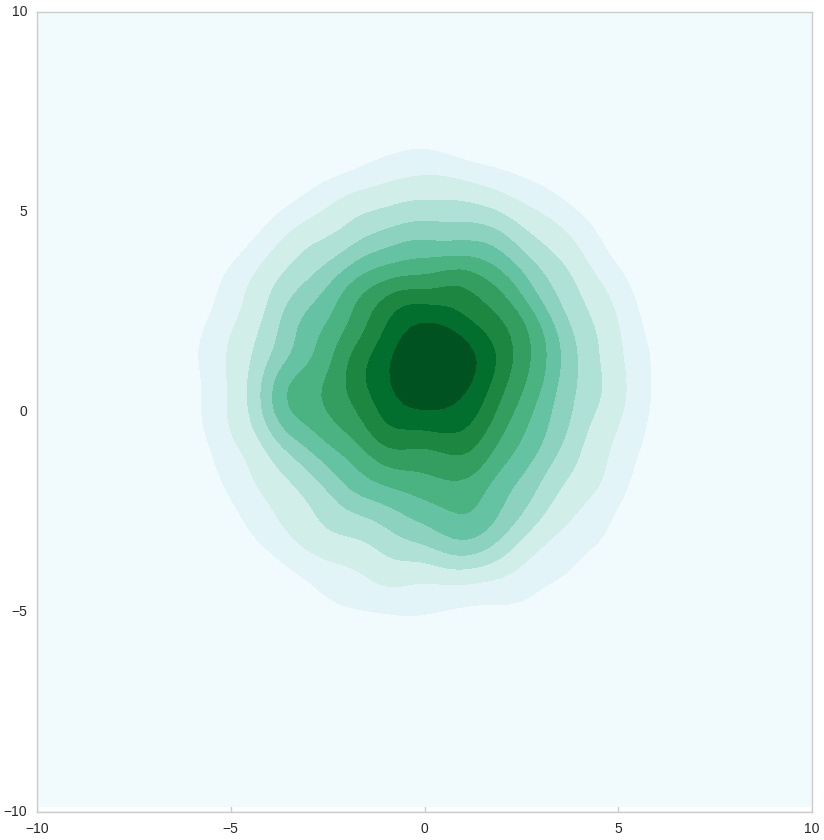}} \\
\subcaptionbox{After 2 LIS modules}{\includegraphics[width=0.45\textwidth]{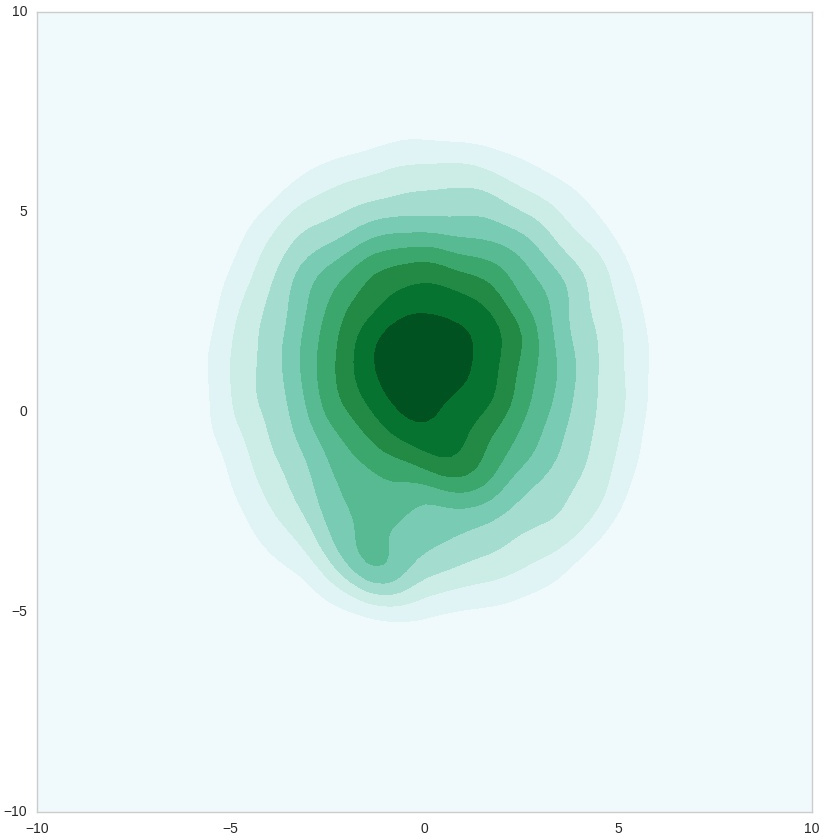}} &
\subcaptionbox{After 3 LIS modules}{\includegraphics[width=0.45\textwidth]{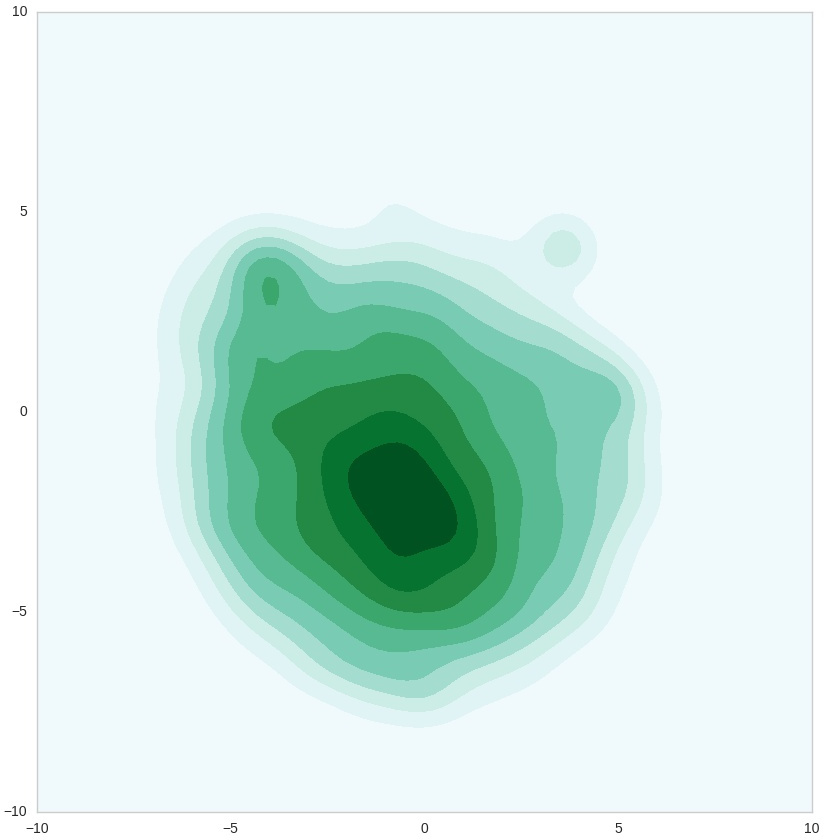}}
\end{tabular}
\caption[t-SNE embeddings]{
    Overview of the distributions of noise vectors, before and after applying LIS modules, based on a CelebA model with three LIS modules (see experiments section). The images show kernel density plots of the t-SNE~\protect\cite{t-SNE} embeddings of 17.5k randomly sampled noise vectors.
    The first image is generated from the initial input. Each of these vectors consists of 256 values sampled from a standard normal distribution. The vectors form a high-dimensional spherical cloud with more density in the center. After embedding, the cloud is converted into a circular shape. With every LIS module (and especially the last one), the embedding deviates more and more from that shape. This indicates that the LIS modules do not only change mean values and standard deviations of the components, but instead seem to prefer some areas of the input space.
}
\label{img:appendix-tsne}
\end{figure}

% -------------------------------------

\clearpage
\section{Probability distributions of noise vector components}
\label{sec:appendix-densities}

\begin{figure}[h]
\begin{tabular}{cc}
\subcaptionbox{Component 1}{\includegraphics[width=0.45\textwidth]{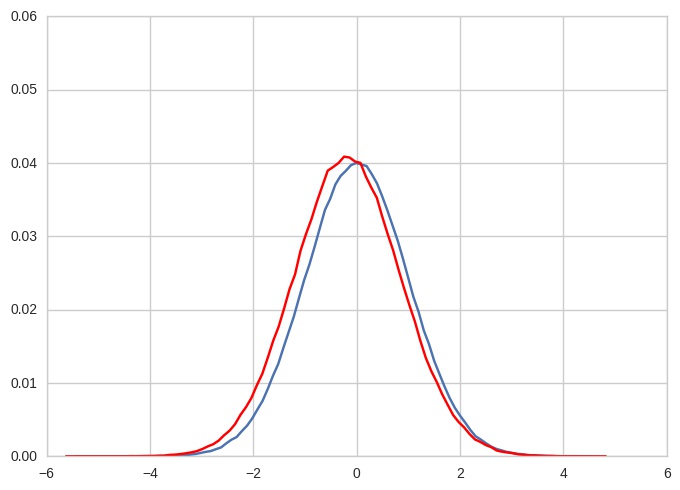}} &
\subcaptionbox{Component 2}{\includegraphics[width=0.45\textwidth]{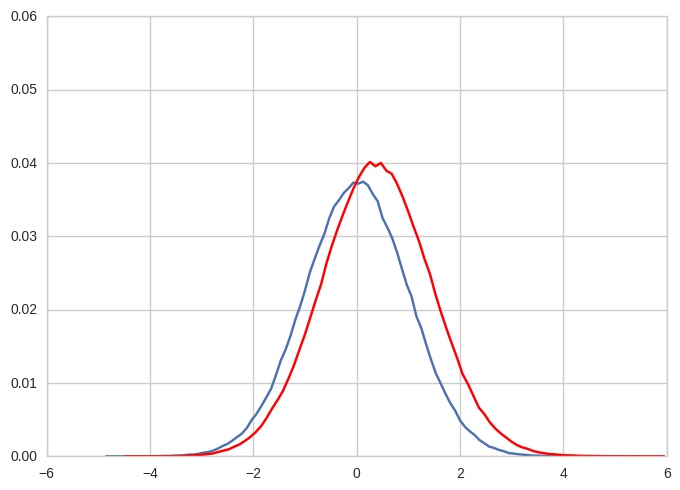}} \\
\subcaptionbox{Component 3}{\includegraphics[width=0.45\textwidth]{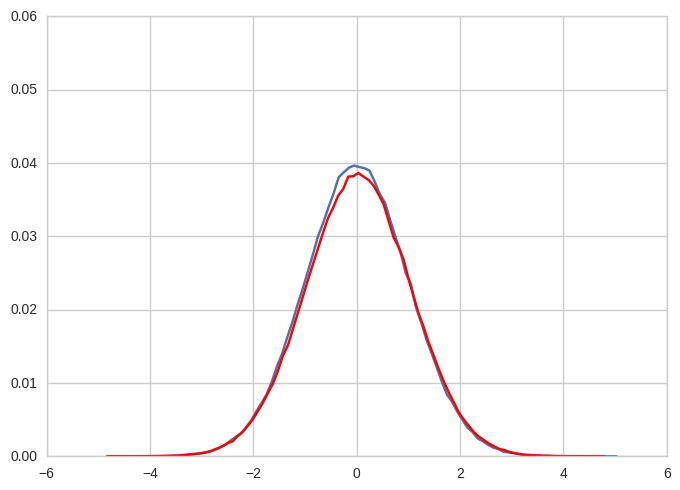}} &
\subcaptionbox{Component 4}{\includegraphics[width=0.45\textwidth]{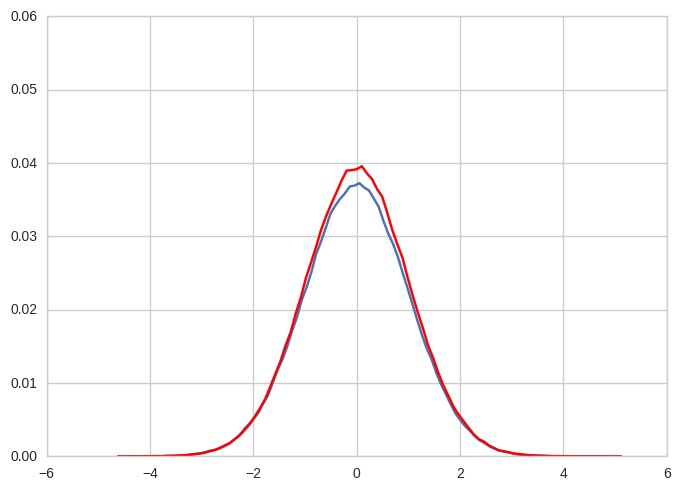}} \\
\subcaptionbox{Component 5}{\includegraphics[width=0.45\textwidth]{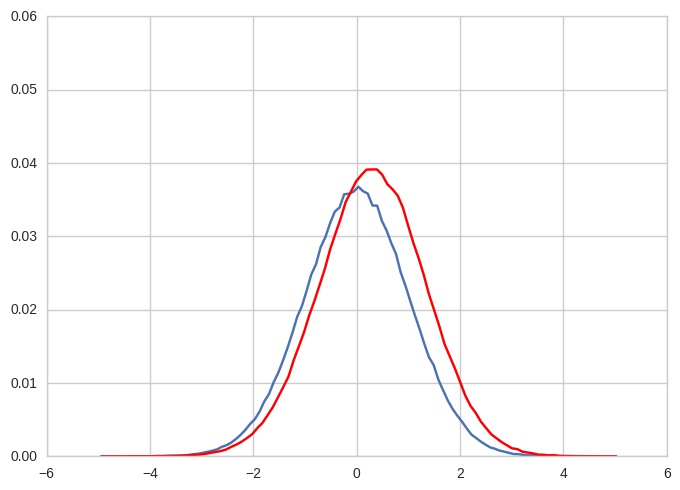}} &
\subcaptionbox{Component 6}{\includegraphics[width=0.45\textwidth]{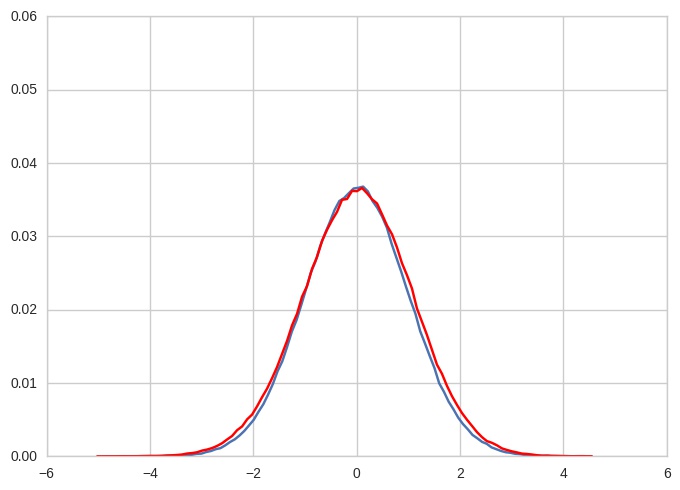}}
\end{tabular}
\caption{
    Histogram-based approximations of the probability distributions of the first six (of 256) components of noise vectors, before (blue) and after (red) they were changed by a model via LIS modules. The approximations are based on 500k randomly sampled noise vectors. The model that manipulated the vectors was trained on CelebA and contained three LIS modules (see experiments section). The red line references the outputs of the last of these modules. As the noise vector's values were sampled from the standard normal distribution, the initial distributions (blue line) are roughly zero-mean with unit-variance (slight deviations are possible due to the approximation). The model changes these distributions, though in most cases it seems to focus on the mean and not the standard deviation. Interestingly, it does not seem to add "bumps" anywhere and instead keeps the bell-shape. We made similar observations for the other components.
}
\label{img:appendix-components}
\end{figure}

% ------------------------------------------

\clearpage
\section{Loss curves of G-LIS}
\label{sec:appendix-loss-curves}

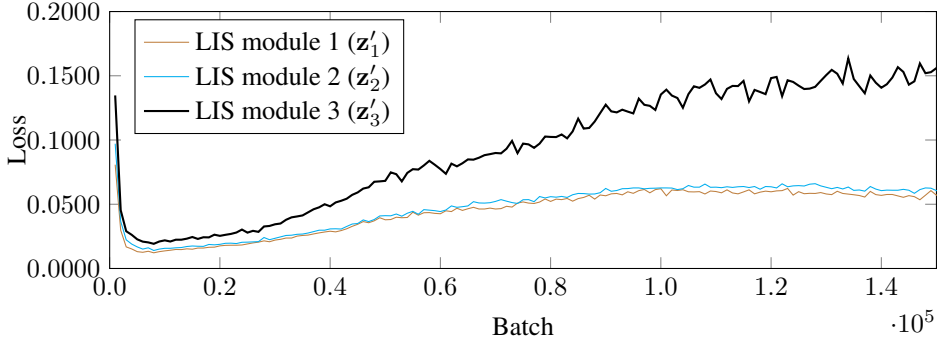
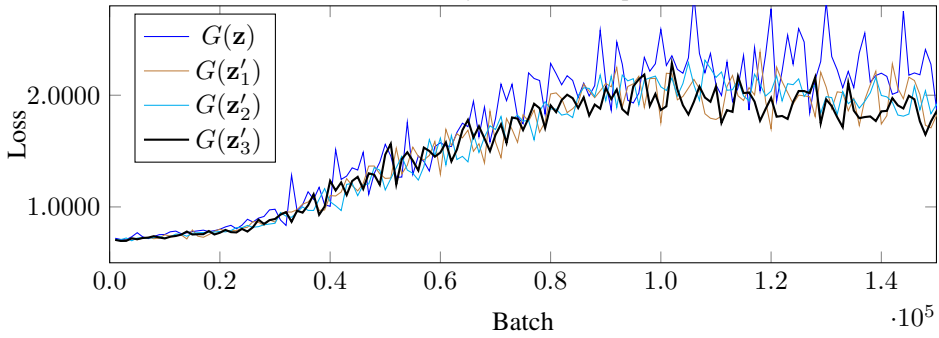
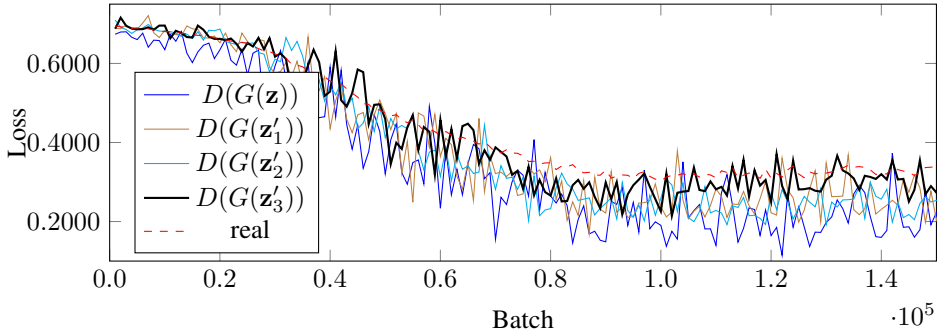
\begin{figure}[h]
\begin{subfigure}{1\textwidth}
%
% ------------
% Similarity constraint
% ------------
\begin{tikzpicture}
\begin{axis}[
    width=0.9\textwidth,
    height=5.0cm,
    xlabel={Batch},
    ylabel={Loss},
    xmin=0,
    xmax=150000,
    ymin=0,
    ymax=0.2,
    y tick label style={
        /pgf/number format/.cd,
            fixed,
            fixed zerofill,
            precision=4,
        /tikz/.cd
    },
    x tick label style={
        /pgf/number format/.cd,
            fixed,
            fixed zerofill,
            precision=1,
        /tikz/.cd
    },
    %mark repeat={50}
    %mark=none
    no markers,
    legend pos=north west
]
\addplot [brown] table [x=__index, y=loss-r-mix_train-r0, col sep=comma] {data/glis_exp01_history6.csv};
\addlegendentry{LIS module 1 ($\mathbf{z}_1'$)};
\addplot [cyan] table [x=__index, y=loss-r-mix_train-r1, col sep=comma] {data/glis_exp01_history6.csv};
\addlegendentry{LIS module 2 ($\mathbf{z}_2'$)};
\addplot [black,thick] table [x=__index, y=loss-r-mix_train-r2, col sep=comma] {data/glis_exp01_history6.csv};
\addlegendentry{LIS module 3 ($\mathbf{z}_3'$)};
\end{axis}
\end{tikzpicture}
\caption{Loss of similarity constraint $\mathcal{L}_R$ per LIS module}
\end{subfigure}
%
% ------------
% Generator
% ------------
\begin{subfigure}{1\textwidth}
\begin{tikzpicture}
\begin{axis}[
    width=0.9\textwidth,
    height=5.0cm,
    xlabel={Batch},
    ylabel={Loss},
    xmin=0,
    xmax=150000,
    ymin=0.5,
    ymax=2.8,
    y tick label style={
        /pgf/number format/.cd,
            fixed,
            fixed zerofill,
            precision=4,
        /tikz/.cd
    },
    x tick label style={
        /pgf/number format/.cd,
            fixed,
            fixed zerofill,
            precision=1,
        /tikz/.cd
    },
    %mark repeat={50}
    %mark=none
    no markers,
    legend pos=north west
]
\addplot [blue] table [x=__index, y=loss-g-mix_train-g0, col sep=comma] {data/glis_exp01_history6.csv};
\addlegendentry{$G(\mathbf{z})$};
\addplot [brown] table [x=__index, y=loss-g-mix_train-g1, col sep=comma] {data/glis_exp01_history6.csv};
\addlegendentry{$G(\mathbf{z}_1')$};
\addplot [cyan] table [x=__index, y=loss-g-mix_train-g2, col sep=comma] {data/glis_exp01_history6.csv};
\addlegendentry{$G(\mathbf{z}_2')$};
\addplot [black,thick] table [x=__index, y=loss-g-mix_train-g3, col sep=comma] {data/glis_exp01_history6.csv};
\addlegendentry{$G(\mathbf{z}_3')$};
\end{axis}
\end{tikzpicture}
\caption{Loss of generator}
\end{subfigure}
%
% ------------
% Discriminator
% ------------
\begin{subfigure}{1\textwidth}
\begin{tikzpicture}
\begin{axis}[
    width=0.9\textwidth,
    height=5.0cm,
    xlabel={Batch},
    ylabel={Loss},
    xmin=0,
    xmax=150000,
    ymin=0.10,
    ymax=0.75,
    y tick label style={
        /pgf/number format/.cd,
            fixed,
            fixed zerofill,
            precision=4,
        /tikz/.cd
    },
    x tick label style={
        /pgf/number format/.cd,
            fixed,
            fixed zerofill,
            precision=1,
        /tikz/.cd
    },
    %mark repeat={50}
    %mark=none
    no markers,
    legend pos=south west
]
\addplot [blue] table [x=__index, y=loss-d-mix_train-d-fake0, col sep=comma] {data/glis_exp01_history6.csv};
\addlegendentry{$D(G(\mathbf{z}))$};
\addplot [brown] table [x=__index, y=loss-d-mix_train-d-fake1, col sep=comma] {data/glis_exp01_history6.csv};
\addlegendentry{$D(G(\mathbf{z}_1'))$};
\addplot [cyan] table [x=__index, y=loss-d-mix_train-d-fake2, col sep=comma] {data/glis_exp01_history6.csv};
\addlegendentry{$D(G(\mathbf{z}_2'))$};
\addplot [black,thick] table [x=__index, y=loss-d-mix_train-d-fake3, col sep=comma] {data/glis_exp01_history6.csv};
\addlegendentry{$D(G(\mathbf{z}_3'))$};
\addplot [dashed, red] table [x=__index, y=loss-d-mix_train-d-real, col sep=comma] {data/glis_exp01_history6.csv};
\addlegendentry{real};
\end{axis}
\end{tikzpicture}
\caption{Loss of discriminator}
\end{subfigure}
\caption{
    Example loss curves of a training run of G-LIS with three LIS modules on CelebA.
    (a) Losses of the similarity constraint $\mathcal{L}_R$ for each of the three modules. The third one violates the constraint significantly more than the others. As the discriminator becomes better at detecting good and bad images, the loss of the generator grows and the LIS modules become more active in order to prevent errors early on.
    (b) Loss of the generator. It is highest for images corresponding to the original noise vector $\mathbf{z}$ and tends to be lowest for the noise vectors generated by the last LIS module.
    (c) Loss of the discriminator, for real images and fake images corresponding to $\mathbf{z}$ as well as all outputs of the three LIS modules. The images generated based on the last LIS module's outputs tend to result in high loss of the discriminator, i.~e. it tends to mistake them more often for real images.
}
\label{appendix-loss-curves}
\end{figure}

% ----------------------------------------

\clearpage
\section{Network Architectures}
\label{sec:appendix-architectures}

Below are the network architectures used in the previous experiments. These are mostly identical to the ones in~\cite{WN}. The generator is here outlined with one LIS module (as in G-LIS). More than one module was used in some experiments. Generators without such a module simply skip the layer. Each LIS module contains two fully connected layers of which only the first one is normalized and has an activation function (listed here under "hyperparameters"). See~\cite{WN} for explanations regarding WN (weight normalization) and TPReLUs.

\begin{table}[!hp]
\centering
\begin{tabular}{c|c|c}
Layer & Layer type & Hyperparameters \\ \hline
input & $\mathbf{z}$ & 256 components sampled from $\mathcal{N}(0,1)$ \\
1 & LIS module & 2x256 Neurons (first FC layer: WN, TPReLU) \\
2 & Fully Connected & 12800 Neurons, WN, TPReLU \\
3 & Reshape & (5, 5, 512) \\ 
4 & Convolution & 256 filters, 4x4, fractionally strided, WN, TPReLU \\
5 & Convolution & 128 filters, 4x4, fractionally strided, WN, TPReLU \\
6 & Convolution & 64 filters, 4x4, fractionally strided, WN, TPReLU \\
7 & Convolution & 3 filters, 4x4, fractionally strided, affine WN, Sigmoid \\
\end{tabular}
\caption{
    Generator (G) for 80x80 images (with one LIS module).
}
\label{tbl:G-80x80-architecture}
\end{table}

\begin{table}[!hp]
\centering
\begin{tabular}{c|c|c}
Layer & Layer type & Hyperparameters \\ \hline
input & $\mathbf{x}$ & (80, 80, 3) \\
1 & Convolution & 64 filters, 4x4, stride 2, WN, TPreLU \\
2 & Convolution & 128 filters, 4x4, stride 2, WN, TPreLU \\
3 & Convolution & 256 filters, 4x4, stride 2, WN, TPreLU \\
4 & Convolution & 512 filters, 4x4, stride 2, WN, TPreLU \\
5 & Convolution & 1 filter, 5x5, stride 1, affine WN, Sigmoid \\
\end{tabular}
\caption{
    Discriminator (D) for 80x80 images.
}
\label{tbl:D-80x80-architecture}
\end{table}

\begin{table}[!hp]
\centering
\begin{tabular}{c|c|c}
Layer & Layer type & Hyperparameters \\ \hline
input & $\mathbf{x}$ & (80, 80, 3) \\
1 & Convolution & 32 filters, 4x4, stride 2, WN, TPreLU \\
2 & Convolution & 64 filters, 4x4, stride 2, WN, TPreLU \\
3 & Convolution & 128 filters, 4x4, stride 2, WN, TPreLU \\
4 & Convolution & 256 filters, 4x4, stride 2, WN, TPreLU \\
5 & Convolution & 256 filters, 5x5, stride 1, affine WN, no activation \\
\end{tabular}
\caption{
    Reverser (R) for 80x80 images.
}
\label{tbl:R-80x80-architecture}
\end{table}

\clearpage

\begin{table}[!hp]
\centering
\begin{tabular}{c|c|c}
Layer & Layer type & Hyperparameters \\ \hline
input & $\mathbf{z}$ & 256 components sampled from $\mathcal{N}(0,1)$ \\
1 & LIS module & 2x256 Neurons (first FC layer: WN, TPReLU) \\
2 & Fully Connected & 25600 Neurons, WN, TPReLU \\
3 & Reshape & (5, 5, 1024) \\ 
4 & Convolution & 512 filters, 4x4, fractionally strided, WN, TPReLU \\
5 & Convolution & 256 filters, 4x4, fractionally strided, WN, TPReLU \\
6 & Convolution & 128 filters, 4x4, fractionally strided, WN, TPReLU \\
7 & Convolution & 64 filters, 4x4, fractionally strided, WN, TPReLU \\
8 & Convolution & 3 filters, 4x4, fractionally strided, affine WN, Sigmoid \\
\end{tabular}
\caption{
    Generator (G) for 160x160 images (with one LIS module).
}
\label{tbl:G-160x160-architecture}
\end{table}

\begin{table}[!hp]
\centering
\begin{tabular}{c|c|c}
Layer & Layer type & Hyperparameters \\ \hline
input & $\mathbf{x}$ & (160, 160, 3) \\
1 & Convolution & 64 filters, 4x4, stride 2, WN, TPreLU \\
2 & Convolution & 128 filters, 4x4, stride 2, WN, TPreLU \\
3 & Convolution & 256 filters, 4x4, stride 2, WN, TPreLU \\
4 & Convolution & 512 filters, 4x4, stride 2, WN, TPreLU \\
5 & Convolution & 1024 filters, 4x4, stride 2, WN, TPreLU \\
6 & Convolution & 1 filter, 5x5, stride 1, affine WN, Sigmoid \\
\end{tabular}
\caption{
    Discriminator (D) for 160x160 images.
}
\label{tbl:D-160x160-architecture}
\end{table}

% --------------------------

\clearpage

\section{Algorithms}
\label{sec:appendix-algorithms}

Below are the algorithms used for \textit{R-separate} and \textit{R-iterative}, which are mostly copies from~\cite{GAN}. \textit{G-LIS} follows the standard GAN algorithm, with the minor exceptions of (a) backpropagating the losses associated with the similarity constraints $\mathcal{L}_R$ through the LIS modules and (b) randomly stopping the execution of the LIS modules before the $i$-th module with probability $0.5^{N_R-i}$. As these are only small changes, the algorithm for \textit{G-LIS} is not listed here.

\begin{algorithm}
\caption[Algorithm R-separate]{
    Algorithm to train R-separate, adapted from~\protect\cite{GAN}. G is the generator, D the discriminator, R the reverser. $m$ is the number of examples per batch. $N_z$ is the number of components per noise vector.
}
\label{alg:R-seperate}
\begin{algorithmic}[1]
\For{number of training iterations of G/D}
    \State Sample minibatch of $m$ noise examples $\{\mathbf{z}^{(i)}, \dots, \mathbf{z}^{(m)}\}$ from noise prior $p_g(\mathbf{z})$.
    \State Sample minibatch of $m$ examples $\{\mathbf{x}^{(1)}, \dots, \mathbf{x}^{(m)}\}$ from data generating distribution.
    \State Update the discriminator by ascending its stochastic gradient:
    \begin{equation*}
        \nabla_{\theta_d} \frac{1}{m} \sum_{i=1}^{m}\left[ log D\left(\mathbf{x}^{(i)}\right) + \mathrm{log}\left(1-D\left(G\left(\mathbf{z}^{(i)}\right)\right)\right)\right].
    \end{equation*}
    \State Sample minibatch of $m$ noise examples $\{\mathbf{z}^{(i)}, \dots, \mathbf{z}^{(m)}\}$ noise prior $p_g(\mathbf{z})$.
    \State Update the generator by descending its stochastic gradient:
    \begin{equation*}
        \nabla_{\theta_g} \frac{1}{m} \sum_{i=1}^{m}\mathrm{log}\left(1-D\left(G\left(\mathbf{z}^{(i)}\right)\right)\right).
    \end{equation*}
\EndFor

\For{number of training iterations of R}
    \State Sample minibatch of $m$ noise examples $\{\mathbf{z}^{(i)}, \dots, \mathbf{z}^{(m)}\}$ from noise prior $p_g(\mathbf{z})$.
    \State Update the reverser by descending its stochastic gradient:
    \begin{equation*}
        \nabla_{\theta_r} \frac{1}{m} \sum_{i=1}^{m} \sum_{j=1}^{N_z} \frac{1}{2}\left(\mathbf{z}_j^{(i)} - R(G(\mathbf{z}^{(i)}))_j \right)^2.
    \end{equation*}
\EndFor
\end{algorithmic}
\end{algorithm}

\begin{algorithm}
\caption[Algorithm R-iterative]{
    Algorithm to train R-iterative, adapted from~\protect\cite{GAN}. G is the generator, D the discriminator, R the reverser. $N_R$ is the number of iterations to execute R+G in order to generate images (G is always executed once at the start). $m$ is the number of examples per batch. $N_z$ is the number of components per noise vector. $\lambda_R$ is a strength weighting for the similarity constraint.
}
\label{algo:R-iterative}
\begin{algorithmic}[1]
\For{number of training iterations}
    \For{$t \in \left[0, .., N_R\right]$}
        %\Comment{Load noise vectors or approximate them via R from the last iteration's images.}
        \If{first iteration}
            \State Sample minibatch of $m$ noise examples $\{\mathbf{z}^{(i)}, \dots, \mathbf{z}^{(m)}\}$ from noise prior $p_g(\mathbf{z})$.
            \State Save these vectors as $\mathbf{\dot{z}}$.
        \Else
            \State Approximate this iteration's noise vectors using the reverser
            \begin{equation*}
                R(\mathbf{\dot{x}^{(i)}})
            \end{equation*}
            \State Save these vectors as $\mathbf{\dot{z}}$.
        \EndIf
        
        %\Comment{Determine whether to start training at this iteration}
        \If{previous iteration was trained}
            \State Set $p$ to $1.0$.
        \Else
            \State Set $p$ to $\frac{1+t}{1+N_R}$.
        \EndIf
        
        %\Comment{Train, if coinflip hits.}
        \If{a coinflip with probability $p$ returns $1$}
            \State Sample minibatch of $m$ examples $\{\mathbf{x}^{(1)}, \dots, \mathbf{x}^{(m)}\}$ from data generating distribution.
            \State Update the discriminator by ascending its stochastic gradient:
            \begin{equation*}
                \nabla_{\theta_d} \frac{1}{m} \sum_{i=1}^{m}\left[ log D\left(\mathbf{x}^{(i)}\right) + \mathrm{log}\left(1-D\left(G\left(\mathbf{\dot{z}}^{(i)}\right)\right)\right)\right].
            \end{equation*}
            \State Sample minibatch of $m$ noise examples $\{\mathbf{z}^{(i)}, \dots, \mathbf{z}^{(m)}\}$ noise prior $p_g(\mathbf{z})$.
            \State Update the generator by descending its stochastic gradient:
            \begin{equation*}
                \nabla_{\theta_g} \frac{1}{m} \sum_{i=1}^{m}\mathrm{log}\left(1-D\left(G\left(\mathbf{\dot{z}}^{(i)}\right)\right)\right).
            \end{equation*}
            
            \If{the reverser was used to approximate $\mathbf{\dot{z}}$}
                \State Update the reverser by descending its stochastic gradient:
                \begin{align*}
                    \nabla_{\theta_r} & \lambda_R^{(1+t)} \frac{1}{m} \sum_{i=1}^{m} \sum_{j=1}^{N_z} \frac{1}{2}\left(\mathbf{z}_j^{(i)} - \mathbf{\dot{z}}^{(i)}_j \right)^2 \\
                    & + (1-\lambda_R^{(1+t)}) \frac{1}{m} \sum_{i=1}^{m}\mathrm{log}\left(1-D\left(G\left(\mathbf{\dot{z}}^{(i)}\right)\right)\right).
                \end{align*}
            \EndIf
        \EndIf
        
        %\Comment{Save images for the next iteration, so that R can approximate noise vectors.}
        \State Save generated images as $\mathbf{\dot{x}}$.
    \EndFor
\EndFor
\end{algorithmic}
\end{algorithm}

\end{appendices}

\end{document}